# Automated data processing and feature engineering for deep learning and big data applications: a survey

Alhassan Mumuni[1]* and Fuseini Mumuni[2]

**Abstract**—Modern approach to artificial intelligence (AI) aims to design algorithms that learn directly from data. This approach has achieved impressive results and has contributed significantly to the progress of AI, particularly in the sphere of supervised deep learning. It has also simplified the design of machine learning systems as the learning process is highly automated. However, not all data processing tasks in conventional deep learning pipelines have been automated. In most cases data has to be manually collected, preprocessed and further extended through data augmentation before they can be effective for training. Recently, special techniques for automating these tasks have emerged. The automation of data processing tasks is driven by the need to utilize large volumes of complex, heterogeneous data for machine learning and big data applications. Today, end-to-end automated data processing systems based on automated machine learning (AutoML) techniques are capable of taking raw data and transforming them into useful features for Big Data tasks by automating all intermediate processing stages. In this work, we present a thorough review of approaches for automating data processing tasks in deep learning pipelines, including automated data preprocessing–e.g., data cleaning, labeling, missing data imputation, and categorical data encoding–as well as data augmentation (including synthetic data generation using generative AI methods) and feature engineering–specifically, automated feature extraction, feature construction and feature selection. In addition to automating specific data processing tasks, we discuss the use of AutoML methods and tools to simultaneously optimize all stages of the machine learning pipeline.

**Index Terms**—AutoML, automated data pre-processing, automated data processing, generative AI, automated feature engineering, automated machine learning.

## 1 INTRODUCTION

### 1.1 Background

Today, many problems are solved using machine learning (ML) methods. Deep learning approaches are generally preferred to traditional machine learning techniques for data-intensive tasks because of their ability to automatically extract useful features from data and perform low-level data processing. While deep learning methods have performed incredibly well, in the era of Big Data they have been shown to lack the power necessary to handle large volumes of complex heterogeneous data [1], [2]. Moreover, there are many pertinent issues (e.g., bias [3], the presence of anomalies [4], [5] and missing data points [6]) that require specific workarounds in deep learning pipelines. Consequently, additional steps are often needed to handle these complex data processing problems and to improve the predictive performance and reliability of machine learning models in big data applications. In big data analytics, data processing is primarily aimed at simplifying the representation of data to reveal meaningful patterns, interrelationships and important trends. This is important in decision support systems, where human decision making is enhanced by insights gained as a result of processing large quantities of relevant data. Figure 1 shows a simplified workflow of a typical big data processing system.

The machine learning model design process itself is a time-consuming process requiring extensive domain knowledge. To train a model for image classification, for example, the developer would usually accomplish the task through the following steps: (1) Collect and fine-tune image data, (2) choose a suitable ML algorithm, (3) define acceptable parameter and hyperparameter settings, (4) train the selected model on the training data, (5) assess the performance of the resulting model on test data, and (6) repeat the procedure from steps 3-6 until satisfactory performance is achieved. With this process, domain experts are needed to collect relevant data, carry out initial data preparation and perform additional processing and feature engineering to ensure that the resulting data is suitable for the specific machine learning task.

The basic workflow of these processing steps is shown in Figure 2. In this paper, we explore various methods for automating data processing tasks for deep learning and big data applications. In the context of this survey, we broadly divide data processing tasks into three main subtasks: preprocessing, data augmentation and feature engineering (i.e., feature processing) functions.

### 1.2 Related works

Approaches for automating data acquisition and processing functions have received significant attention recently. Interest in these methods is largely driven by the need to leverage

. [1]*Alhassan Mumuni: Department of Electrical and Electronics Engineering, Cape Coast Technical University, Cape Coast, Ghana.*
*\*Corresponding author, E-mail: alhassan.mumuni@cctu.edu.gh*
. [2]*Fuseini Mumuni: University of Mines and Technology, UMaT, Tarkwa, Ghana. E-mail: fmumuni@umat.edu.gh*

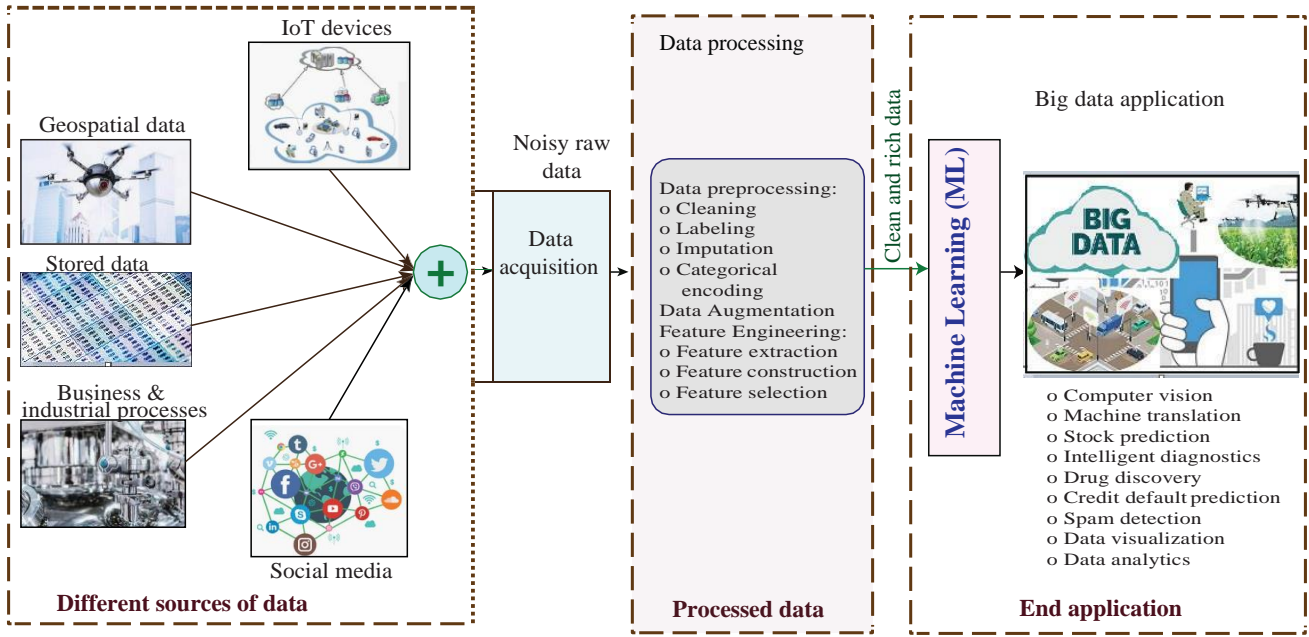

Figure 1. Simplified representation of the basic components and workflow of a typical Big Data application. Typical Big Data applications rely on aggregating multimodal data from several different sources, and then applying suitable techniques to process these raw data to train deep learning models for downstream tasks.

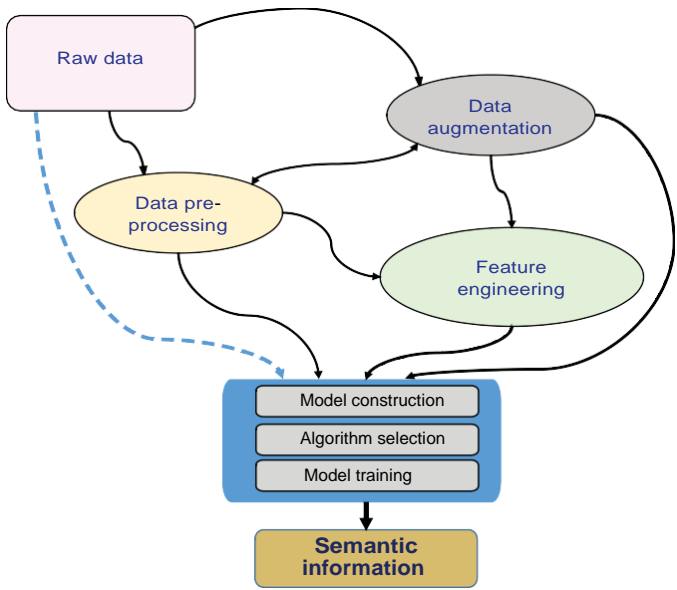

Figure 2. Basic workflow of data preparation and feature engineering in machine learning and Big Data application development. There are a range of data preparation tasks (oval nodes) that may be performed before the data can be used for training. The arrows show the direction of of data processing steps. Note that, depending on the application requirements and the quality and quantity of the input data, some steps can be bypassed.

the enormous amount of data available in various forms from diverse sources for machine learning and big data applications. At the same time, the complexity of machine learning problems has increased drastically while requirements have also become more stringent. Additional factors such as data privacy and ethical issues limit access to data in application settings such as business, healthcare, security and law enforcement. Automated data processing systems provide a way for machine learning systems to process data in situations where privacy issues may restrict access of the raw data to human actors. Indeed, healthcare [7], [8] and business applications are already benefiting immensely from these techniques [9], [10], [11].

Despite the importance and resurgence of automated data processing methods, only a small number of survey works have been authored on these methods. With the exception of a few works such as He et al. [12] and Zöller and Huber [13], who dedicate a small part of their survey to automated data processing methods, most of the published works [13], [14], [15], [16] are focused on ready-to-use, end-to-end automated machine learning (AutoML) tools designed for general purpose applications or deal with the entire pipelines of custom AutoML implementations [7], [16] specific use-cases without particular focus on data processing.

Strikingly, important aspects of low-level data processing tasks, including preprocessing and feature engineering methods have not been discussed sufficiently in the literature. While there exists a large number of extensive surveys on data augmentation methods [17], [18], [19], till date, relatively few surveys [6], [20], [21], [22], [23] have been written on data preprocessing and feature engineering methods. However, the techniques described are mostly based on traditional methods. Our work is motivated by the scarcity of literature specifically focused on current approaches to automating data processing functions in deep learning models, especially data preprocessing and feature engineering tasks.

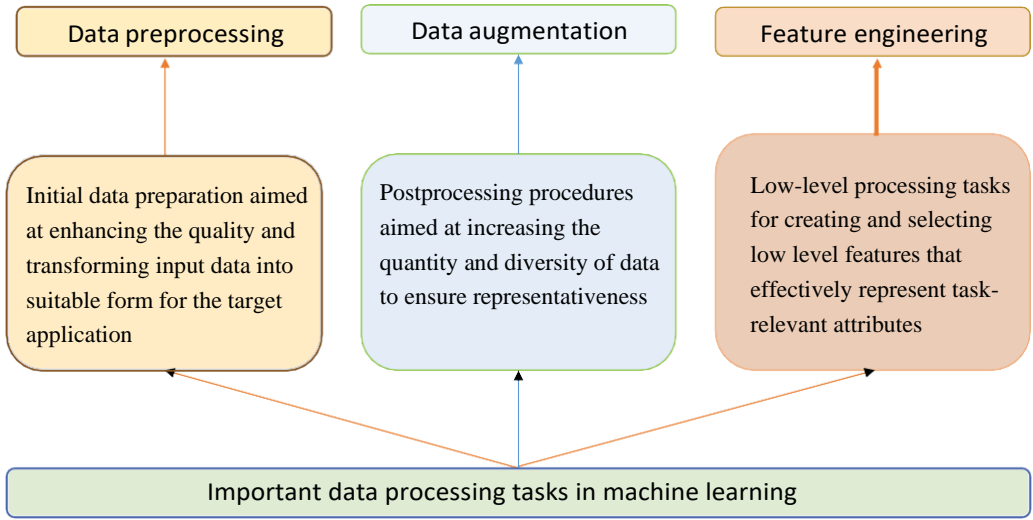

Figure 3. Operational definition of the main data processing tasks discussed in this paper.

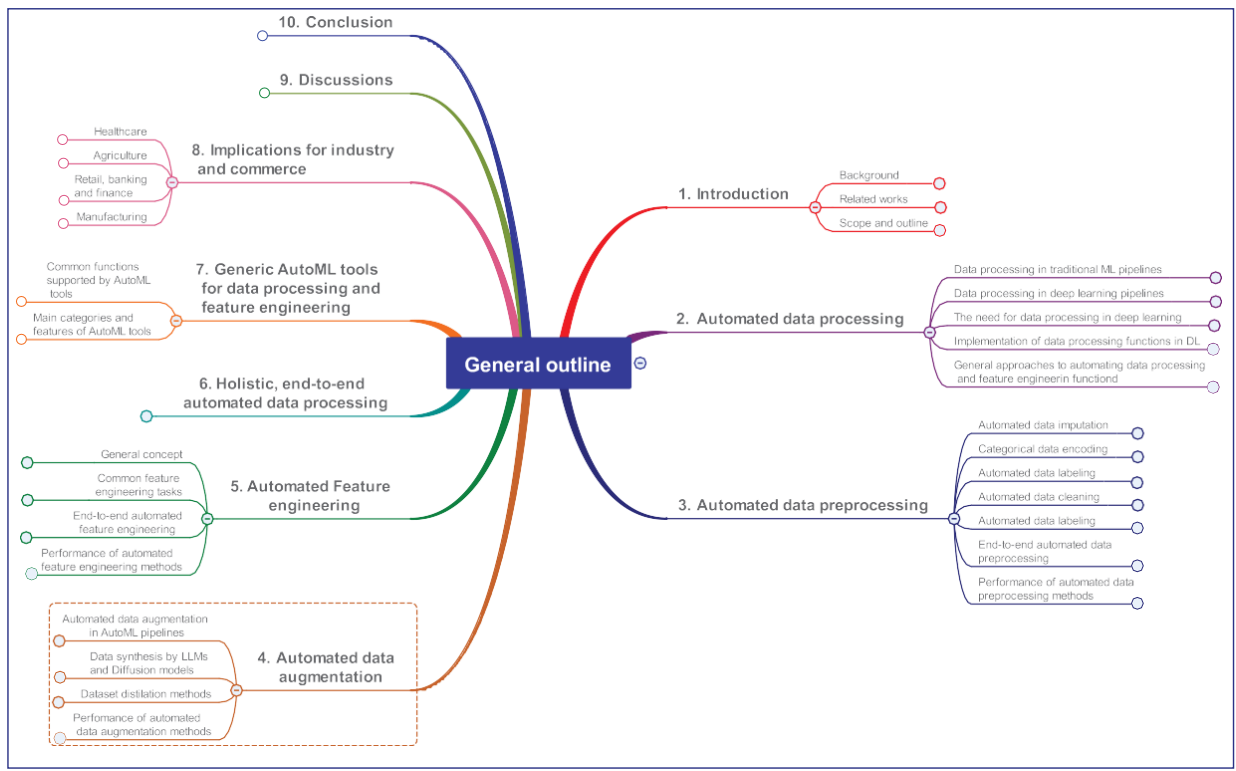

Figure 4. Broad outline and structure of this paper.

## 1.3 Scope and outline of survey

This work focuses on approaches for automating data pre-processing and feature engineering tasks in deep learning pipelines. While several recent surveys [7], [12], [24] have discussed issues on the entire AutoML pipeline –data preparation, feature engineering model generation, algorithm selection, performance evaluation and validation–our work is specifically focused on data preprocessing and feature engineering functions within these pipelines. For completeness, we also discuss data augmentation methods. In this survey, these functions are collectively referred to as data processing functions. The main data processing tasks discussed in this survey are presented in Figure 3.

The rest of the survey is organized as follows. Section **2** provides an overview of the rationale, concepts and methods for automating data processing tasks. In Section 3, automated data preprocessing techniques are covered. Data augmentation is treated in Section 4. Section 5 covers feature engineering methods. In Section 6, we discuss end-to-end workflows for data processing in deep learning pipelines. The focus is on approaches to performing all the different data processing tasks, including data preprocessing, aug-mentation and feature engineering, simultaneously in an end-to-end manner using a single machine learning framework. Section 7 summarizes the main features, application domains and categories of generic end-to-end AutoML tools used for data processing. In Section 8, we present an overview of the implications of automated data processing techniques for industry and commerce, and provide an extensive discussion of pertinent issues and future prospects of automated data processing approaches in Section 9. We conclude in Section 10. Figure 4 (a) presents a broad outline of the this paper.

## 2 AUTOMATED DATA PROCESSING IN MACHINE LEARNING PIPELINES

### 2.1 Data processing in traditional machine learning pipelines

The machine learning problem typically involves the following sequence of steps: data collection, data preparation and preprocessing, data augmentation, extraction of useful features from the data, construction of new features and pruning of the generated feature set (i.e., feature selection), selection of a suitable machine learning model, and optimization of model hyperparameters. Traditional machine learning approaches [25], [26] require all of these stages of the machine learning process to be performed manually and as standalone processes (Figure 5).

This approach presents a number of difficulties that make it extremely challenging to scale up machine learning models. For instance, traditional machine learning pipelines typically incorporate hand-crafted feature extractors to mine useful low-level features from the training data for subsequent processing by the machine learning model. These feature extractors are basically case-specific algorithms aimed at finding the best set of features for the target task. Feature extractors, once designed, are typically fixed – i.e., in the process of learning, they cannot be modified automatically to work differently in response to new conditions. To adjust their operation, the algorithm may have to be redesigned and deployed anew.

### 2.2 Data processing in deep learning pipelines

With modern deep learning models based on neural network frameworks [27], [28], data processing tasks are usually implemented within the machine learning framework and trained end-to-end (Figure 6). Deep learning approaches employ neural networks that are generally able to extract useful low-level information from raw data without the need for prior knowledge or additional processing. Information processing in deep learning pipelines roughly follows the information processing properties of the brains of biological systems, where different semantic features are extracted using a large and functionally diverse network of complex neural structures. Several hierarchical layers of artificial neural networks extract different types of features at different semantic levels. In the case of computer vision, for example, shallower layers extract useful information relating to more basic, low-level visual concepts such as lines, object contours and shapes. Deeper layers sequentially increase the abstraction level of extracted information, mining higher and semantically meaningful information such as composite shapes and whole objects.

Unlike traditional machine learning pipelines where fixed feature extractors are responsible for information processing, in deep learning pipelines, during training, the synaptic elements that extract useful features use tunable parameters to adjust their operation on the fly based on feedback on the performance of the network on the target task. Thus, in the learning process the model automatically adjusts its way of feature extraction and low level data processing in response to performance results.

### 2.3 The need for additional data processing in deep learning models

#### 2.3.1 The need for data preprocessing

Despite the high degree of automation of data processing in deep learning models, their performance has been shown to be highly dependent on the quality of training data [29], [30], [31]. However, in many practical scenarios, the qualitative properties of raw data are not often consistent with the requirements of the target application or model [2], [32]. Consequently, data preprocessing has become an essential task in the deep learning application development process. Preprocessing includes initial data preparation tasks such as data normalization, cleaning (removal of outliers and inconsistent data points), encoding of categorical values, imputation of missing values and labeling.

#### 2.3.2 The need for data augmentation

In many Big Data applications, the collected data may not be sufficient or representative enough for the target task. In such situations, after initial preprocessing, data augmentation is used to increase the quantity and variability of the training data by generating new samples through manippulating the existing data. This is the most popular approach to enhancing the generalization power of deep learning models. Data augmentation has also been used to addressed problems such as bias [33], imbalanced data [34] and domain shift [35].

#### 2.3.3 The need for feature engineering

While machine learning models constructed on the basis of deep neural architectures can automatically learn useful features [36], some application settings still required explicit feature processing to guarantee satisfactory performance. This is particularly so for complex and data-scarce applications such as fault diagnosis in machinery (e.g., in [37]), online fraud detection (e.g., [38]), credit risk evaluation [39] and industrial applications [40]. Indeed, it has been shown in [41] that employing dedicated feature extractors in deep learning networks provides more useful feature representations of the input data that results in significant performance improvements.

Also, for big data applications, the data is generally very large, heterogeneous, complex and noisy, leading to the need to handle very large volumes of features, often with redundant and duplicated samples. This can be harmful to performance [3], [42]. This problem can be mitigated by feature engineering methods, which aim to find compact feature representations for the data. Another common shortcoming of features naturally learned in the natural process of training deep neural networks is that they are usually not interpretable. This can be a major limitation in mission-critical application settings such as medical domains [43]. Feature engineering is also useful for tasks where training data is insufficient to adequately encode the desired information [44].

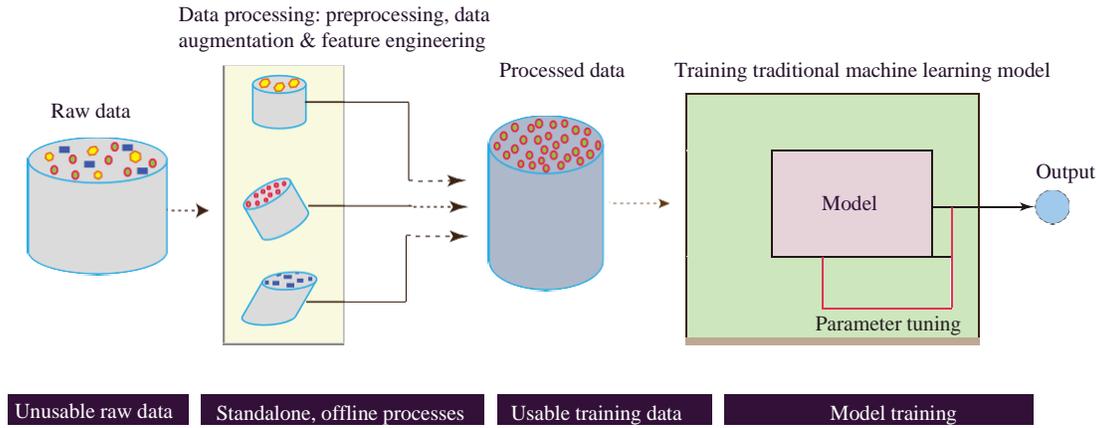

Figure 5. Data processing in traditional machine learning. All processing steps are implemented manually as standalone tasks.

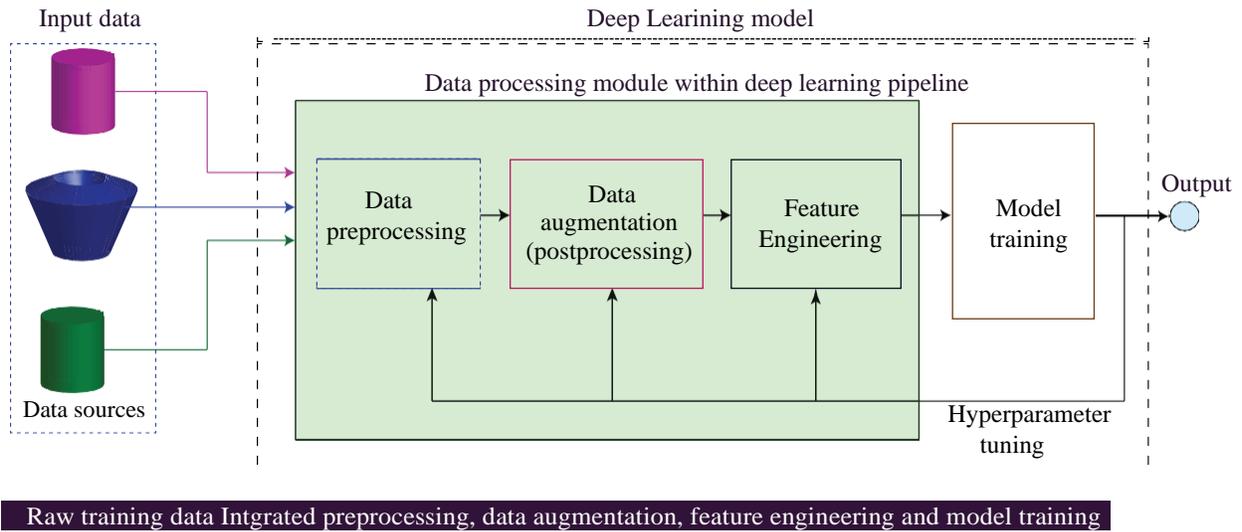

Figure 6. Data processing in deep learning: All processing tasks–preprocessing, data augmentation and feature engineering functions–as well as model generation and training are realized as a unified process.

### 2.4 Implementation of data processing functions in deep learning pipelines

Unlike in shallow learning where data processing is performed offline as a standalone process, modern deep learning favors an end-to-end learning paradigm that aims at realizing all functions in an integrated manner using a single model. This means in deep learning, additional data processing, when needed, is usually integrated into the machine learning process and implemented as a unified process (Figure 6). The fact that this process is carried out online means that no additional samples are stored. This has the advantage of increasing the volume and diversity of training data without a corresponding increase in the amount of storage requirements. Instead of applying explicitly designed transformation operations to create additional data, many new works (e.g., [45], [46], [47], [48], [49], [50]) have devised special neural network architectures to learn the relevant transformations directly from the training data. In this case, the data augmentation tasks become an integral part of the overall learning process.

### 2.5 General approaches to automating data processing and feature engineering tasks

A wide range of techniques are used to automate data processing tasks. These techniques have different strengths and limitations related to complexity, flexibility and predictive performance. While the crudest approaches to automating data processing tasks involve using hand-crafted routines, more advanced methods [51], [52], [53] utilize learned mechanisms to perform processing. The most advanced methods, however, employ automated machine learning (AutoML) techniques [12], [54], [55], [56] to automate data processing functions. These methods can perform all the underlying data preprocessing tasks end-to-end without delegating intermediate tasks to human developers.

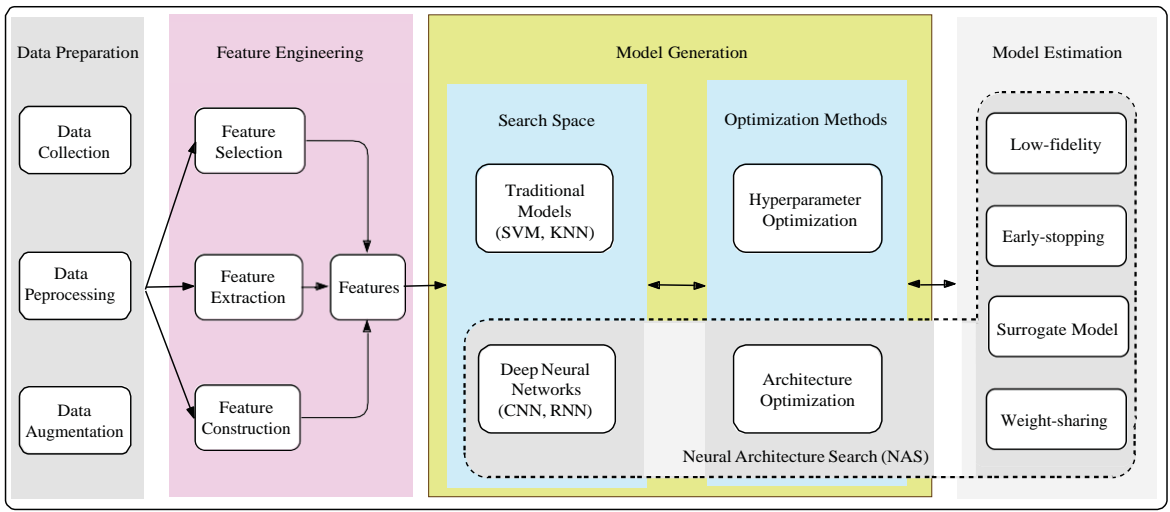

Figure 7. General representation of AutoML pipeline. It incorporates deep learning and traditional machine learning models in its design. In the training process, optimization algorithms select the best model structure and hyperparameter setting for the specific task. The original illustration, from [12], has been modified slightly to reflect the context of the current paper.

A typical AutoML pipeline for end-to-end data processing is shown in Figure 7. The first part of the pipeline is concerned with data processing tasks. These include preprocessing, data augmentation and feature engineering. The second part of the AutoML pipeline performs functions such as model construction, hyperparameter setting, algorithm selection and result evaluation and validation. Additional functions such as result visualization (e.g., in [57], [58]) may also be automated. In some cases (e.g., [59]), the automated machine learning problem involves composing an optimum model architecture to solve the machine learning task. Approaches for automating model structure construction belong to a class of automated machine learning methods called neural architecture search or NAS [60]. We describe the main features and functions of generic automated machine learning systems in Section 7.

## 3 AUTOMATED DATA PREPROCESSING

*The concept of data preprocessing*

Data preprocessing functions consist of a set of basic operations that transform raw data into a form that is useful for the machine learning model. Important preprocessing subtasks include data cleaning [61], [62], labeling or re-labeling [53], [63], [63], [64], categorical encoding [65], [66], [67] and imputation of missing data [68], [69], [70]. Figure 8 depicts the main categories of data preprocessing tasks and the set of common problems they commonly tackle.

*Automated preprocessing*

Given a dataset $D$, the automated preprocessing problem can be defined as a task involving the automatic selection and application of a set of basic preprocessing operations $P_i \leq : i = 1, 2, 3, ..., n$ on $D$ such that the predictive performance on a target task is maximized. This is a challenging task because, besides determining which transformations are appropriate and to what extent to transform the data, the ML algorithm needs to also apply the operations in the correct order [71]. The most advanced automated preprocessing methods [72] typically take as input a target dataset and a set of primitive transformation operations and their corresponding hypeparameters–parameters that define certain details about the model, including the strength of transformations and the order of their application. Optimization algorithms such as reinforcement learning (RL), gradien descent (GD), Bayesian optimization (BO) or evolutionary computational algorithms (ECA) are then employed to search for the best combination of basic preprocessing operations.

*Aproaches to automated preprocessing*

Data preprocessing can be automated to varying degrees. The degree of automation depends on how preprocessing operations and the underlying models are implemented. The most basic automation still requires the use of traditional, hard-coded techniques while the most advanced methods utilize AutoML techniques. The general approaches for automating data processing tasks are summarized in Table 2.

Generally, all automated preprocessing approaches use some form of machine learning to define and/or select an ensemble of data preprocessing operators an/or deep learning pipelines from a set of possible options that maximize performance. We discuss in detail specific approaches for automating some of the most important preprocessing tasks in the next subsections.

One major difficulty with data preprocessing is that not only are the processing methods dependent on data type, they are also highly sensitive to the machine learning model [73]. This means for different models and tasks, the application of the same preprocessing operations on the same type of input data can often lead to vastly different outcomes. Applying all possible preprocessing operations irrespective of model or data type is a highly combinatorial problem and impractical to carry out in applications dealing with very large data. Consequently, many fully-automated preprocessing methods rely on first identifying the input data and model types and applying context-specific preprocessing operations. Unfortunately, in some situations it is not often possible to determine beforehand what operations could be effective for particular models or data types. To solve this problem, approaches (e.g., [74]) have been proposed to jointly estimate the performance of primitive preprocessing operations so as to apply them in a data- and model-dependent way.

Large-scale AutoML frameworks such as AutoGluon-Tabular [75] incorporate a two-level preprocessing scheme, with the first level implementing model-agnostic preprocessing operations while the second one generally focuses on model-specific preprocessing. These AutoML models typically combine many different optimization algorithms and pipelines with different structures that define data preprocessing blocks and machine learning model blocks within a single framework. The composite nature of this class of models makes generic AutoML tools flexible and multipurpose, and allows them to solve a wide range of problems.

### 3.1 Automated data imputation

Handling missing data is a common task that is encountered when developing machine learning models [76]. The problem of missing data arises from various causes associated with data acquisition, including lost data entries, inability to collect certain data, or from situations where some data is simply not accessible for various reasons such as privacy concerns. Missing data can also result from incorrect measurements and human errors, especially in data systems relying on manual entry. Traditional data imputation methods (e.g., [77], [78]) rely on analytically formulated statistical techniques. The methods first identify missing data using knowledge about the statistical properties of the observed data. Missing values are then computed and incorporated into the training set. In contrast, automated imputation techniques [79], [80] discover and correct issues with missing data without encoding explicit mechanisms for doing so.

Generative modeling approaches have emerged as an important class of methods for automated missing data imputation (e.g., in [51], [81], [82], [83], [84], [85], [86]). These methods use learned mechanisms to encode the correct distribution of "normal" data. The learned knowledge about the underlying data distribution can then be used to help detect instances of missing data and perform imputation in an end-to-end manner. Yoon et al., for instance, propose a so-called generative adversarial imputation network (GAIN) [81] that utilizes a generative adversarial network (GAN) to model data distribution for identification and imputation of missing data. Gondara and Wang [82] perform data imputation with the the help of denoising autoencoders. Generative techniques that utilize variational autoncoders are also common (e.g., [52], [87], [88]).

More recently, methods based on AutoML techniques [62], [68], [79], [89] have emerged as the most important class of methods for automating data imputation. Teague [68], for example, develops a tool based on AutoML technique to automatically impute missing tabular data. Hyperimpute [89], implements missing data imputation by utilizing an AutoML framework that optimizes multiple candidate machine learning models using different search algorithms. Its basic structure is shown in Figure 9. The approach can realize a more generalized and adaptive imputation process than methods that rely on optimizing a single machine learning model using one search technique. In addition to these dedicated data imputation models, most popular automated machine learning tools such as AutoSklearn [90], Scikit-learn [91] and Azure Databricks AutoML [92] aim to, among other things, perform automatic imputation in an end-to-end manner.

It has been recognized that fully automated methods of data imputation can be detrimental to performance in some situations [93]. While machine learning techniques can generally detect situations of missing data points, determining the correct data input can sometimes require some form of high-level understanding of the underlying context which machine learning models often lack. To address this limitation, some authors adopt human-in-the-loop approaches (e.g., [80], [94], [95]). Bilal et al. in [80], for example, implement a mechanism to automatically detect missing data and prompts the user, through a data visualization interface, to accept or reject suggested candidate data modifications.

### 3.2 Categorical data encoding

Categorical data encoding is aimed at converting categorical data into numerical information. This pre-processing step is often necessary because deep learning models typically work well with numerical data but are unable to handle categorical data. However, many real-world datasets often include categorical variables. Traditional approaches to converting categorical variables into numerical representations rely on techniques such as binary encoding, ordinal encoding, label encoding, one-hot encoding and target encoding [65].

Currently, it is challenging to completely automate entire process of categorical data encoding. Unlike pre-processing functions such as data cleaning and imputation, it is difficult to define primitive operations that can be applied by machine learning models to convert between categorical and numerical data. Because of this limitation, many approaches, including state-of-the-art AutoML models such as TPOT [96] and Autokeras [97] currently implement encoding manually. Some AutoML-based categorical encoding methods [54], [98], [99] employ traditional encoding schemes within deep learning networks. The approach in [54] incorporates custom implementations of traditional categorical data encoding and other pre-processing functions within a generic AutoML framework. It employs a tree-based ensemble technique to evaluate and select the encoded elements. Similarly, H2O AutoML framework [100], implements categorical data encoding using OneHotEncoding library–an implementation of the traditional one-hot encoding method.

Modern AutoML tools that attempt to automate this pre-processing step (e.g., LightAutoML [101], AutoGluon-Tabular [75], Auto-sklearn [90] and Pharm-AutoML) mostly require some form of human supervision. For example, Auto-sklearn includes a so-called label encoder - a categorical data to integer conversion algorithm, but it requires user input to define the underlying categories before performing the necessary operations.

While many approaches rely on using traditional techniques within automated learning frameworks to perform categorical to numerical data conversion, some recent works suggest encoding categorical data with the help of learned word embeddings. The authors in [102], for instance, propose a word embedding method based on deep neural network model to learn useful features for encoding categorical data. BERT-Sort [99] employs a learned embedding approach that exploits semantic relationships among ordinal data elements to perform categorical encoding. The basic idea is to learn the context of textual information with the help of a pre-trained language model. This is then used to encode input data according to the high-level semantic context of the underlying text.

| DATA CLEANING | DATA LABELING & RE-LABELING | DATA IMPUTATION | CATEGORICAL DATA ENCODING |
|---|---|---|---|
| Invalid data<br>• Ordering errors<br>• Noisy or corrupted data<br>• Redundant data<br>• Irrelevant data<br>• Wrong data types or categories<br>• Inconsistent formats<br>• Overall low quality data | • Unlabeled data<br>• Tagging errors (for crowd-sourced data)<br>• Incorrect labels<br>• Incorrect category nomenclature<br>• Typographical errors in labels<br>• Domain shift | • Missing data points<br>• Missing attributes<br>• Incomplete data<br>• Missing values<br>• Missing keys<br>• Missing meta data<br>• Incomplete metadata<br>• Insufficient data<br>• Balance imbalanced data | • Need to convert ordinal data to numerical<br>• Need to convert nominal data to numerical<br>• Need to weight data or attributes<br>• Need to ensure compatibility of data across different systems |

Figure 8. Data preprocessing tasks and the set of common problems they typically tackle.

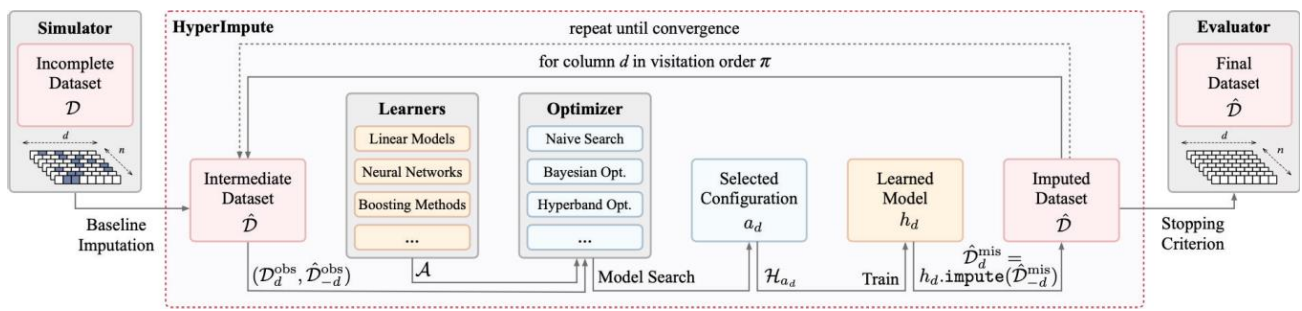

Figure 9. Basic architecture and workflow of HyperImpute, an automated data imputation method proposed in [89]. The approach incorporated different ML models and optimization techniques and uses to find an optimal imputation pipeline.

### 3.3 Automated data cleaning

Data cleaning [103] is aimed at correcting errors in a dataset or eliminating nosy data. The main sources of these errors include ordering and indexing mistakes, incorrect class assignments and inconsistent naming. In machine learning domains such as natural language processing, common errors include typographical, grammatical and syntax errors. Random errors and anomalies can also have significant impact on the performance of time-series prediction. Duplicate or redundant and irrelevant data may also be present in the training dataset and may need to be removed to preserve the fidelity of the data. Data cleaning may also involve removing incomplete–i.e., absence of some necessary attribute(s) required for the target task–incorrect or invalid (e.g., wrong data types) samples. Operations such as normalization, zero centering and scaling are additionally used to ensure that the statistical characteristics of the data are suitable for use in the deep learning model.

Basic data cleaning automation can be realized by implementing high-level wrapper functions around low-level data cleaning routines. CleanTS [62], for instance, develops a set of high-level abstractions and functions at a semantic level on top of a wide-ranging suit of low-level library functions to automate data cleaning tasks in time-series applications.

A recent trend in data cleaning is to use machine learning approach as a way to identify discrepancies and errors and apply appropriate corrections. Learning-based data cleaning methods (e.g., [95], [104], [105], [106]) optimize parameters associated with data cleaning operations based on the performance the underlying network on downstream tasks. Learn2Clean [104], for example, employs a Q-learning algorithm to learn the best data cleaning operations. BoostClean [105] relies on a learning approach that utilizes gradient boosting to automate the data cleaning process.

Meta-learning techniques (e.g., [107], [108]) have been employed to extend the scope and enhance the adaptability of data cleaning systems. This allows them to be applied as generic data cleaning models to process a wide variety of datasets for many different machine learning tasks. End-to-end AutomML-based data cleaning pipelines [62], [98] have been very proposed as stand-alone tools for generic data cleaning purposes. For instance, Shende et al. in [62] develop a generic time series data cleaning method and accompanying toolset that tackles problems with outliers, duplicates, as well as inconsistent data types and formats.

## 3.4 Automated data labeling

Labeling, or annotation, is an important task in data preparation and in supervised machine learning in general. It is arguably the most tedious job in many machine learning tasks. Unfortunately, in many domains, it is also one of the most difficult processes to automate. At present, most annotations are generated either manually or semi-automatically, that is, using algorithms to generate possible label proposals which are subsequently refined and validated by human agents (e.g., in [109], [110], [111], [112], [113]). Only a few works in computer vision [114], [115] and medical image analysis [110], [116] domains have attempted carrying out the labeling process in a fully automatic manner.

Automated labeling techniques in computer vision and medical image analysis applications [110], [112], [117] typically employ classification, object object or semantic segmentation detection models to first identify instances before generating labels for them. Ince et al. [112] employ a standard object detector model as the basis for annotations but requires user validation of the generated labels. Zhang et al. [117] use metric learning approach to match unlabeled images with existing category information and then associate the unlabeled images with labels of relevant image categories. The basic principle of the approach is illustrated in Figure 10.

Another common method is to exploit the rich textual descriptions naturally associated with certain types of data to obtain annotations for unlabeled samples. This is particularly common in medical imaging domains [110], [118], where images are often associated with pathology reports that provide relevant descriptions of the corresponding image content that can be leveraged though training. Zhang et al. [110], for instance, proposed an automatic annotation mechanism for thyroid nodules in ultrasound scans using a natural language processing (NLP) model in combination with fixed rules (ontologies) to extract nodule information from medical pathology reports.

Even though a variety of machine-guided automated labeling [12], [53], [119], [120], [121] techniques have been proposed, they are generally limited to simple labeling tasks targeting very narrow application settings. It is currently challenging to extend them to more complex or general tasks. The main difficulties include the need for context awareness [115], [122], lexical complexity [118], semantic ambiguity [123] and label subjectivity [124].

## 3.5 End-to-end automated data pre-processing

While a number of pre-processing methods are designed to specifically perform a single preprocessing task (e.g., in [84], [89]), many works (e.g., [61], [61], [80], [126]) tend to simultaneously implement multiple pre-processing functions. For example, Auto-Prep [80]) performs automatic missing data imputation, data type detection and deletion of duplicates, categorical data encoding as well as feature scaling. AutoDC [61] performs outlier detection and elimination, label correction, edge case selection as well as data augmentation.

As we have already mentioned, most of the pre-processing methods based on AutoML frameworks generally aim to automate the entire machine learning pipeline, including data processing as well as model structure construction algorithm selection functions. Because of their generic focus, they are able to handle only basic pre-processing tasks automatically. More complex pre-processing steps often require human guidance to accomplish. Models such TPOT and Autokeras, for example, perform only missing data detection. While human-in-the-machine-learning-loop has been suggested [80], [94] as a viable solution, other researchers [129] suggest performing human and machine learning pre-processing independently and combining the results for better and reliable performance.

Some new approaches such as AutoData [53], DataAssist [125], BioAutoMATED [127], Atlantic [54] and DiffPrep [126] are designed as dedicated pre-processing plug-ins that can be integrated into standard AutoML frameworks. In this way, they can leverage the hyperparameter and pipeline optimization capabilities of large-scale AutoML models to optimize pre-processing functions. These models are generally aimed at addressing the limitations of generic methods and tools by developing specialized modules that can be used to perform specific data preparation tasks within general-purpose AutoML pipelines. AutoData [53] (Figure 11), for instance, implements a dedicated data processing module that can be used within standard AutoML frameworks as an end-to-end mechanism for performing many data pre-processing and and augmentation tasks, including data acquisition, labeling, cleaning and augmentation. It uses reinforcement learning method to search for suitable data from diverse external sources. It first takes a user-defined input specifying the required data type and task, as well as the overall performance objective. Given this input, the reinforcement learning algorithm searches for the appropriate data from external repositories and feeds the results to an AutoML framework (e.g., Auto-WEKA or Auto-Keras) which then constructs a model and evaluates the performance on the given task and provide feedback for refining the search results. Table 2 the mail approaches for automating various pre-processing tasks. These reflect different degrees of automation–from the most basic (i.e., manual) to fully automated pre-processing (based on AutoML).

## 3.6 Performance of automated data preprocessing methods

Despite the proven effectiveness and the proliferation of data preprocessing methods, there is a general lack of quantitative performance results for different data preprocessing methods. The scarcity of comparative performance data in the literature can be attributed to several factors. Firstly, data preprocessing methods often vary widely in their implementation and application; there are hardly any standard techniques for performing specific preprocessing tasks. In particular, there are no standard algorithms for

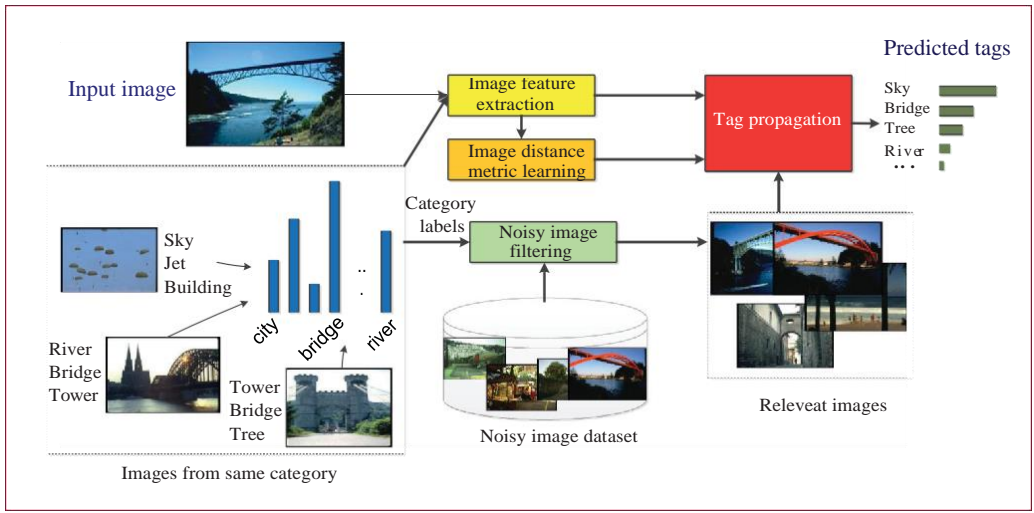

Figure 10. Automatic image annotation by the Zhang et al. [117]: The model first extracts and combines both hand-crafted and deeply learned visual features for both sets of images (i.e., labeled categories and unlabeled images). Metric learning is then used to estimate the similarity of unlabeled images and the known (i.e., labeled) categories. Next, the labels of relevant images from the labeled categories are transferred to the unlabeled instances based on their similarity scores. Information about the labeled categories is also used to filter noisy input samples to provide clean images for labeling.

Table 1

Summary of important data pre-processing methods. We capture the main pre-processing functions (imputation, cleaning, labeling, and categorical encoding) each of the techniques performs and supported the data types. We also indicate the main machine learning tasks, the classsification of the approaches and the datasets used.

| Work | Data type/ Task | Main pre-processing functions | | | | Approach to automation | Main datasets |
|---|---|---|---|---|---|---|---|
| | | Data Imput. | Data Clean. | Data Label. | Categ. encod. | | |
| scIGANs [51] | Tabular (genomics) | √ | | | | Generative modeling | PMBC 10k, Synthetic |
| MIWAE [87] | Tabular (classification) | √ | | | | Generative modeling | UCI Repository |
| Automunge [68] | Tabular, insurance claims, housing price (regression) | √ | √ | √ | √ | Mostly analytical | Ames Housing |
| Hyperimpute [89] | Tabular (classification) | √ | | | | AutoML | Allstate Claims |
| PLACL [63] | Image (keypoint localization) | | | √ | | Deep learned | UCI MPIL, CUB-200-2011, ATRW, MS-COCO, AnimalPose |
| cleanTS [62] | Tabular (time series) | √ | V | √ | | AutoML | Power, Temperature, CO2 Emission |
| GP-VAE [88] | Image, digitized signals (binary format) | √ | | | | Generative modeling (VAE) | Physionet, Healing MNIST, SPRITES |
| DataAssist [125] | Tabular | √ | √ | √ | √ | Deep learned | Various (Kaggle) |
| Gain [81] | Tabular (classification) | √ | | | | Generative modeling | UCI |
| DiffPrep [126] | Tabular (regression) | √ | √ | | √ | AutoML (gradient-based) | OpenML |
| BERT-Sort [99] | Tabular (regression, classification) | | | | √ | Deep learned | Various (UCI, Kaggle) |
| BioAutoMATED [127] | Sequence-based data (genomics) | | √ | | √ | AutoML | Glycan, Peptide sequences, Ribosome binding |
| Rotom [108] | Text, tabular; Entity matching, NLP | | √ | √ | | Deep learned (meta-learning) | TextCLS, EDT, EM |
| Learn2clean [104] | Text, multimedia (web data) | √ | √ | | | Deep learned (RL) | House Prices, Google Play Store Apps, Users |
| AlphaClean [128] | Tabular | | √ | | | AutoML (greedy search) | LAQ, Physician, Hospital |
| AutoData [53] | Image, text (from the web) | | √ | √ | | Deep learned (RL) | Custom (web data) |
| BoostClean [105] | Tabular (classification) | √ | √ | | | Deep learned | Various (UCI, Kaggle, USCensus, NFL, Titanic, etc.) |

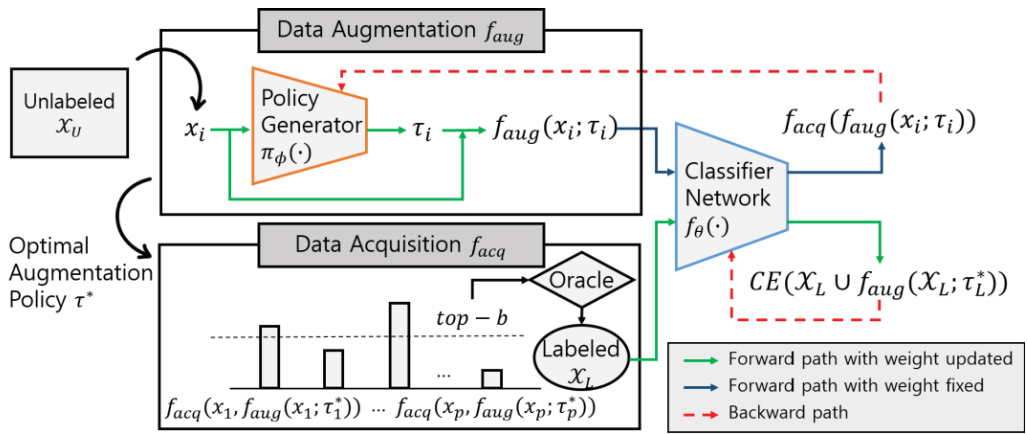

Figure 11. Data Labeling by LADA: The LADA framework proposed in [130] acquires unlabeled training data from the real world. An augmentation policy is then used to carry out initial preprocessing and generate additional data by learning relevant transformation operations. With the aid of a classification network, the optimum policy is determined using the classification loss. Finally, an *oracle* is used to lablel the unlabeled raw data.

Table 2
General approaches for automating data preprocessing tasks. The approaches vary in sophistication, from basic to fully automated methods based on AutoML pipelines.

| Method | Description and main characteristics | Example works |
|--------|--------------------------------------|---------------|
| Basic | Standalone preprocessing, usually manual or with the aid of generic data processing tools (e.g., image processing libraries for manipulating images). This approach is largely manual. | [131], [132] |
| Analytical | Automated processing within deep learning pipelines using explicitly formulated analytical relations. | [45], [49], [133] |
| DL | Deeply learned processing modules within machine learning pipelines. | [51], [83], [130] |
| AutoML | End-to-end processing, model selection and hypaparameter tuning using AutoML. | [53], [61], [126], [54], [68], [134] |

common tasks such as data cleaning, labeling and imputation. Because of the highly varied implementation details, it is difficult to compare performance across different studies.

The second major difficulty is that data preprocessing is rarely accomplished by a single algorithmic operation. Each of the preprocessing procedures (i.e., data imputation, labeling, cleaning, etc.) often involves a set of multiple sequential or parallel operations. Comparing different methods requires defining consistent pipelines, which can be very complex given the many possible combinations of preprocessing procedures.

Another important challenge is that most data preprocessing procedures involve subjective considerations. For example, determining outliers or selecting which data elements to retain can be influenced by domain knowledge and researcher bias. This subjectivity also makes it difficult to obtain fair and objective comparisons.

## 4 AUTOMATED DATA AUGMENTATION

Data augmentation involves creating variations of the original training data by applying context-appropriate transformation operations. For example, in computer vision, these may involve image scaling, shearing, rotation, reflection and blurring operations. Text augmentation methods include synonym replacement, random insertion and deletion. The creation of such variations lead to a more representative data for the given task, and enhances the machine learning model's ability to generalize well on unseen data. Figure 12 summarizes the common data augmentation operations for common machine learning tasks. Typically, for a particular task, data augmentation may involve applying all of these basic operations together with more advanced techniques.

As explained earlier, in most cases data augmentation is achieved by manipulating existing data in such a way as to create adequate variability. Most state-of-the-art methods for automated data augmentation rely on AutoML techniques. There are, however, practical situations where training data does not exist in any form [135], and augmentation methods are required to synthesize novel data from scratch. Generative artificial intelligence techniques have excelled in this area.

Presently, for data manipulation, automation has largely been successful in computer vision and image understanding domains. It has been particularly challenging to extend these methods to time series forecasting and natural language processing domains owing to the difficulty in achieving semantic-preserving transformations without human supervision. Only a very few works (e.g., [136], [137]) have reported successful application of automated data augmentation techniques in natural language processing domains. Recently approaches for generative AI techniques have achieved remarkable results in text generation.

| Machine learning Task | Basic data augmentation methods |
|---|---|
| Image processing and computer vision (image classification, object detection, visual tracking, e.tc.) | • **Rotation**: Rotating images by various angles to simulate different viewpoints.<br>• **Flipping:** Horizontally flipping images to create mirror images<br>• **Cropping**: Randomly cropping images to focus on different parts of the scene<br>• **Zooming**: Applying random zoom to create variations in image scale<br>• **Color JiMering**: Modifying the brightness, contrast, saturation, and hue of images<br>• **Gaussian Noise:** Adding random Gaussian noise to images to simulate real-world variations.<br>• **Blurring:** Applying Gaussian blur or other blurring techniques to images |
| Text recognition & language understanding (machine translation, text-to-voice conversion, etc.) | • **Synonym Replacement**: Replacing words with their synonyms to create variations.<br>• **Random Deletion:** Randomly removing words to change sentence length and structure<br>• **Random Insertion:** Randomly inserting additional words into a sentence to simulate natural language variation<br>• **Text Paraphrasing:** Rephrasing sentences while preserving the original meaning<br>• **Character-Level Perturbations:** Swapping, inserting, or deleting characters to simulate typographical errors or variations in spelling<br>• **Contextualized Word Embedding:** Replacing words with alternatives that are contextually relevant |
| Time series forecasting (e.g., stock price prediction, economic performance forecasting, sales forecasting, etc.) | • **Dynamic Time Warping:** Stretching or compressing the time axis of a time series to induce variations in the temporal scale of the data<br>• **Window Slicing:** Dividing the original time series into overlapping or non-overlapping windows of varying lengths can provide the model with multiple perspectives of the data<br>• **Noise Addition:** Introducing random noise to the time series data<br>• **Data Interpolation**: Filling missing data points using interpolation techniques<br>• **Outlier Injection**: Introducing outliers or anomalies into the time series data can help the model learn how to identify and handle unexpected events.<br>• **Time Shifing:** ShiGing the entire time series or specific portions of it forward or backward in time to create new instances with shiGed patterns. |

Figure 12. Common data augmentation functions for various machine learning tasks. Note that this list is by no means complete.

## 4.1 Automated data augmentation in AutoML pipelines

AutoML has emerged as a powerful augmentation approach for large-scale machine learning and Big Data applications. AutoML-based data augmentation methods generally aimed to automate the process of creating and applying appropriate transformation operations on the training data.In addition to data augmentation, the AutoML models perform a number of additional tasks, such as machine learning model development (e.g., model selection, hyperparameter tuning, performance validation, etc.). In an AutoML framework, a large number of augmentation operations are usually created manually or in an automated way, and the data augmentation task reduces to a simple search for the best transformation operations and their associated hyperparameters. The basic concept of AutoML approach to data augmentation is shown in Figure 13.

### 4.1.1 Generation of augmentation operations

The first step in the augmentation process is to generate diverse transformation operations to manipulate the input data. In image augmentation, for example, the basic transformations are typically standard geometric and photometric image processing functions like rotations, flipping, scaling, shearing, color jittering, solarizaion, noise addition and contrast adjustment. Transformation magnitudes may include scale factors, rotation angles, translation offsets, color intensity and brightness levels, etc. The various transformation operations and their corresponding magnitudes constitute a search space.

The effectiveness of automated augmentation strategies is based largely on the ability to compose a comprehensive set of operations that are representative of the tasks under consideration. The general approach to automating data augmentation tasks involves creating the set of required transformation operations and then applying the created operations on data and algorithmically determining the most useful augmentations using various optimization techniques. We briefly discuss common approaches to creating augmentations as well as techniques for optimizing the augmentation strategies.

*Analytical Methods*

The most basic approach to automating data augmentation is the use of semi-automated or analytical methods to create augmented data. The analytical method involves defining a set of explicit mathematical relations or heuristics to perform required transformations on the training data. These relations can be domain-specific and depend on the characteristics of the given data type. This class of methods generally perform standard augmentation functions such as rotations, shearing, flipping, and color jittering. Many works (e.g., [138], [139], [140], [141], [142] involve defining different types of basic transformations functions to apply, as well as specifying the degree or magnitudes of these operations within an applicable range. While simple, the approaches can be effective for certain tasks [143], [144], they may not be able to account for unknown transformations.

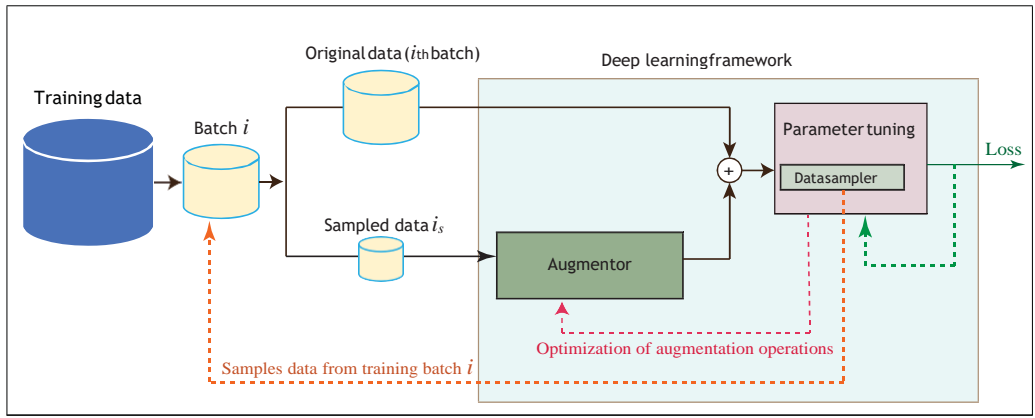

Figure 13. General scheme and process of automated data augmentation methods. A portion of the training set is sampled by a specified algorithmda, ta sampler, for augmentation. The selected sample undergoes a series of transformations by an augmenter sub-module. The augmented sample is then added to the original data batch $i$ for training. The training loss is fed to both the sampler and augmenter to fine-tune their performance. Over time, this results in better sampling and augmentations, leading to overall improved performance.

*Deep-learned transformations*

Some recent works [46], [145], [146], [147], [148] have suggested moving from predefined augmentation operations to learned mechanisms, which allow augmentation operations to be learned from input data rather than relying on explicitly formulated relations. The basic idea is to implement data transformation sub-modules within deep learning pipelines whose parameters and hyperparameters can be leaned and applied on input data in the training process. This method offers a higher degree of automation than approaches based on explicit analytical formulations. They also allow arbitrary or unknown augmentation styles to be achieved. A common technique widely employed for learning transformations is the spatial transformation network (STN) [45] (Figure 14A).

*Generative machine learning techniques*

Another common way to automate the construction of data augmentation operations is based on generative modeling [145], [149], [150], [151]. This approach creates transformed samples without explicitly requiring analytical formulas, or even requiring implicit transformation parameters to be learned. The basic idea is to first learn the distribution of real data by training a discriminator-generator pair on the original dataset. Knowledge about the distribution can then help to generate new but slightly varied samples that simulate data diversity. Thus, in the automated data processing pipeline, the generative model is employed as an intermediate data transformation unit. Its parameters are learned jointly with the CNN model in a bi-level optimization scheme (see Figure 14 B).

Generative machine learning approaches based on GANs are particularly common in computer vision domains, where they enable realistic image samples to be generated with desired visual features. This family of methods is also useful in applications where it is necessary to transfer knowledge from an existing domain to a different domain (e.g., in [152]). Generative methods are also used to align the distributions of data from different sources, enabling deep learning models to perform better when applied to new data that may come from a slightly different distribution.

However, despite the usefulness of the approach, generative models themselves require careful tuning, greatly increasing labor demands, and thus may limit the full benefits of automation.

### 4.1.2 Optimization of augmentation strategies

Optimization techniques are typically employed to search for the best augmentation strategy. The search task involve finding not only relevant transformations but also the optimum levels of transformations. For image augmentation these levels may be rotation angles, translation offsets, saturation values, etc.

Because the search space is inherently complex and discontinuous, many works utilize black-box optimization techniques such as reinforcement learning (e.g., in [138], [140], [140], [142], [153], [154]), Bayesian optimization (e.g., in [142], [155], [156]) and evolutionary computation algorithms (e.g., in [157], [158], [159], [160] to search for good augmentation strategies These techniques have produced impressive results but are noticeably slow when dealing with very large data.

To overcome this limitation, approaches for optimizing the discrete augmentation search space using the concepts of approximation gradients [161] have also been proposed [149], [162], [163], [164]. The use of these concepts makes it possible to develop methods for solving optimization problems for discontinuous functions present in the search space. Approaches based on approximate techniques are generally more efficient than black-box search techniques.

Alternative search spaces providing the ability to find effective augmentation policies without extensive search have also been considered [165], [166]. In some cases (e.g., [167]) these approaches eliminate the need for the search stage altogether. For instance, researchers in [166] contend that it is not necessary to conduct exhaustive search for extensive range of possible augmentations and hyperparameters in a combinatorial manner and, instead, propose to reduce the search space to a linear search space of uniform probabilities and augmentation parameters that can be traversed by a simple grid search. The approaches, despite the significant reduction in compute time, have shown competitive perfor-mance. However, there seems to be little room to further extend their predictive performance.

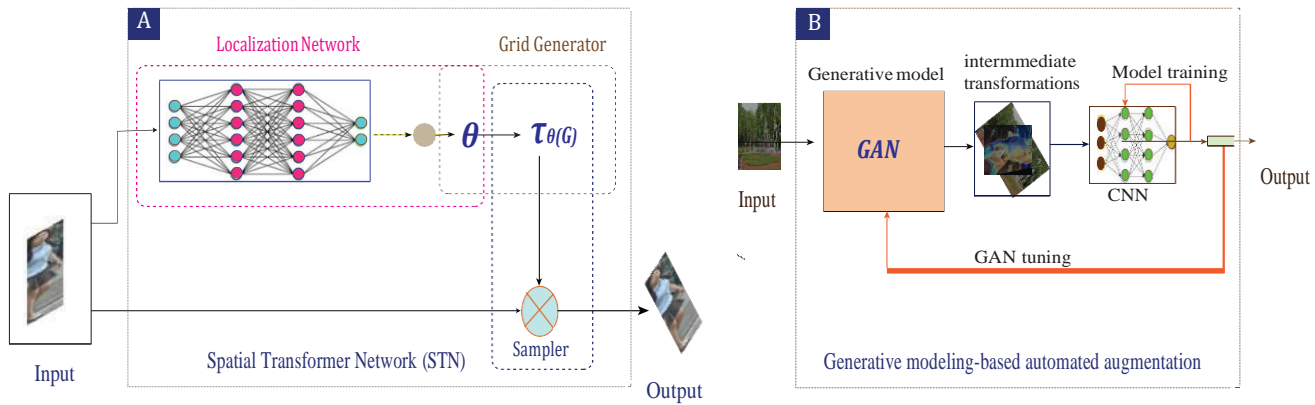

Figure 14. Approaches to generating basic transformations within automated data augmentation frameworks: (A) - learned transformation using spatial transformation network (STN) [45], and (B) - generative-modeling-based synthesis of transformed image data.

## 4.2 Data synthesis by large language models (LLMs) and diffusion models

The automated data augmentation methods covered in Subsection 4.1.1 utilize generative modeling techniques to perform intermediate data processing in AutoML pipelines. While generative methods can be used to perform data transformation operations in that manner, approaches based on variational autoencoders [168], generative adversarial networks [169], diffusion models [170], autoregressive models [171], [172] and large language models (LLMs) [173] are commonly designed to directly synthesize data by themselves, thereby bypassing all intermediate processing steps. Recent advances in deep generative AI-based data synthesis methods, especially diffusion models and large language models (LLMs), have enabled the possibility of generating clean data from scratch or from noisy data in an end-to-end manner. Data generated this way can be used to augment existing data (e.g., in [174], [175], [176]) or completely replace natural data in situations where datasets are inadequate or are inaccessible for training machine learning models (e.g., in [177], [178]).

Large language models are a special type of deep learning systems based on the concept of transformer [179] and are primarily designed to perform NLP tasks. On the other hand, diffusion models are a particularly versatile class of generative methods that synthesize data in a similar way to generative adversarial networks – a technique that allows deep neural networks to generate data using random noise. They are based on nonequilibrium thermodynamics [180]. Diffusion models are realized by first defining a Markov chain of diffusion steps that gradually increase the random noise component to input data, and then implement a reverse diffusion process to re-create the desired data samples from the input noise (Fig 15).

Generative methods based on LLMs [174], [181] and diffusion models [182], [183], [184] have become very useful in augmenting data for natural language processing and computer vision applications, respectfully. In NLP applications, additional text data commonly generated by prompting LLMs such as ChatGPT [185], LLaMA [186] and BERT [187] to complete input sentences with missing words or phrases so that different variations of the original sentences could be created. For instance, Ubani et al. in [188] propose to manipulate text data with the help of ChatGPT to produce augmented data for training deep learning networks for NLP-based tasks by giving appropriate prompts to generate syntactically different variations of the original sentences. In computer vision applications, generative AI-based data augmentation typically involves manipulating existing images to generate novel styles, poses, background contexts and views. DiffusionCLIP [184] introduces an approach to perform various image transformations using text-guided prompts with a diffusion model developed on the basis of CLIP [189]. The approach achieved photorealistic stylizations of different images in a wide range of novel contexts. Figure 16 shows the result of various image stylizations methods using DiffusionCLIP [184] and two other text-based generative data manipulation models –StyleCLIP [190] and StyleGAN-NADA [191].

While diffusion models and LLMs can both perform data processing– e.g., text manipulation [188] and image quality enhancement (super-resolution) [182], [192] , denoising [183] and styling [193], [194], their most powerful use-case has been the task of automatically generating high quality data from "scratch" [195], [196]. Modern generative AI techniques based on diffusion models can be used to synthesize many different types of data – including images [197], [198], videos [199], [200], [201], [202], audio [203], [204], time series [205] and tabular data [206], [207], [208] –using only text prompts without additional input data.

Language models, when trained on pairs of data (e.g., images) and corresponding descriptive texts (e.g., captions), can learn the association between the textual descriptions and the underlying data. The trained model can then be used to generate desired images based on descriptive text inputs. In this regard, large language models and diffusion models play a complementary role in the data generation process. Indeed, the popularity of large-scale diffusion models has been driven largely by the development of powerful generative tools that generate data based on intuitive text prompts enabled by LLM techniques. Some of the most prominent data generation models in this category include Dalle-2 [209], Stable Diffusion [210], Glide [193] and Imagen

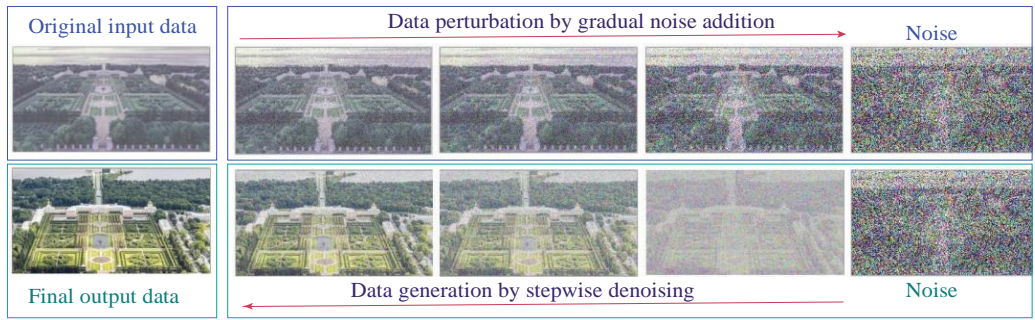

Figure 15. The idea of diffusion models is to learn to generate data by gradually adding random perturbations to the input data, and then perform the reverse process of denoising the data until a clean data output is achieved. Image by authors.

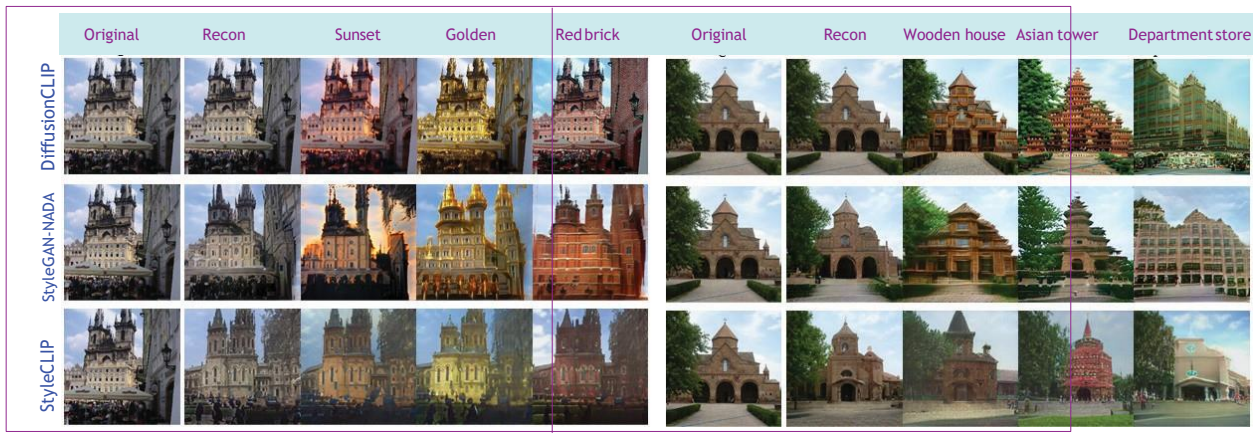

Figure 16. Visual comparison of results of different generative AI-based image stylization methods: DiffusionCLIP [184], StyleCLIP [190] and StyleGAN-NADA [191]. The models attempt to style a church building with respect to different visual contexts. Image is from [184]. We have applied some minor changes in the text fonts to make it clearer.

[211]. Figure 17 depicts photorealistic images generated by different diffusion models based on user's supplied prompts.

### 4.3 Dataset distillation methods

While the approaches discussed in Subsections 4.1 rely on generating extra data to improve machine learning models, an alternative approach, known as dataset distillation [214], [215], seeks to select an informative subset from a large-scale training dataset that retains the important properties and, importantly, the generalization performance of the original dataset. That is, the aim is to produce a leaner dataset that still performs satisfactorily on the target task. The approach can also be used to distill data labels from a large set of labels (e.g., in [216], [217]) or from noisy labels (e.g., in [218]). While traditional approaches to dataset distillation rely on manually engineered procedures, some recent works [219], [220] have proposed mechanisms for automating the process. Although these approaches are well motivated, and can potentially lead to high performing but leaner deep learning models, at present their level of automation is limited. Consequently, their popularity is relatively low.

### 4.4 Performance of automated data augmentation methods

Results from several studies show that models trained on data generated by automated techniques outperform those trained on data generated using manual approaches. In Table 3 we compare the performance of ten state-of-the-art automated data augmentation techniques and ten of the best traditional augmentation methods across several datasets (CIFAR-10 [221], CIFAR-100 [221] and ImageNet [222]). All results are obtained using Wide ResNet (WRN-28-10) [223] backbone trained until convergence -roughly about the same number of epochs (300). This provides a fair ground for comparison of the different methods. In Figure 18 we summarize the comparative performance by averaging the classification scores as well as the percentage improvement over the baseline (i.e., models trained without augmentation). These results convincingly demonstrate the effectiveness of automated data augmentation approaches.

While the performance of automated augmentation techniques is in no doubt, an important challenge is the enormous computational resources typically required for training these models. Also, since the augmentation process is not guided by human intuition, some augmentation operations may unduly alter the original semantic meaning of the data, which may not be observed at training time and can later lead to catastrophic failure. For instance, modifying the label of an image or applying overly aggressive transformations in such a way that they transform its semantic meaning can result in erroneous training signals with potentially harmful implications.

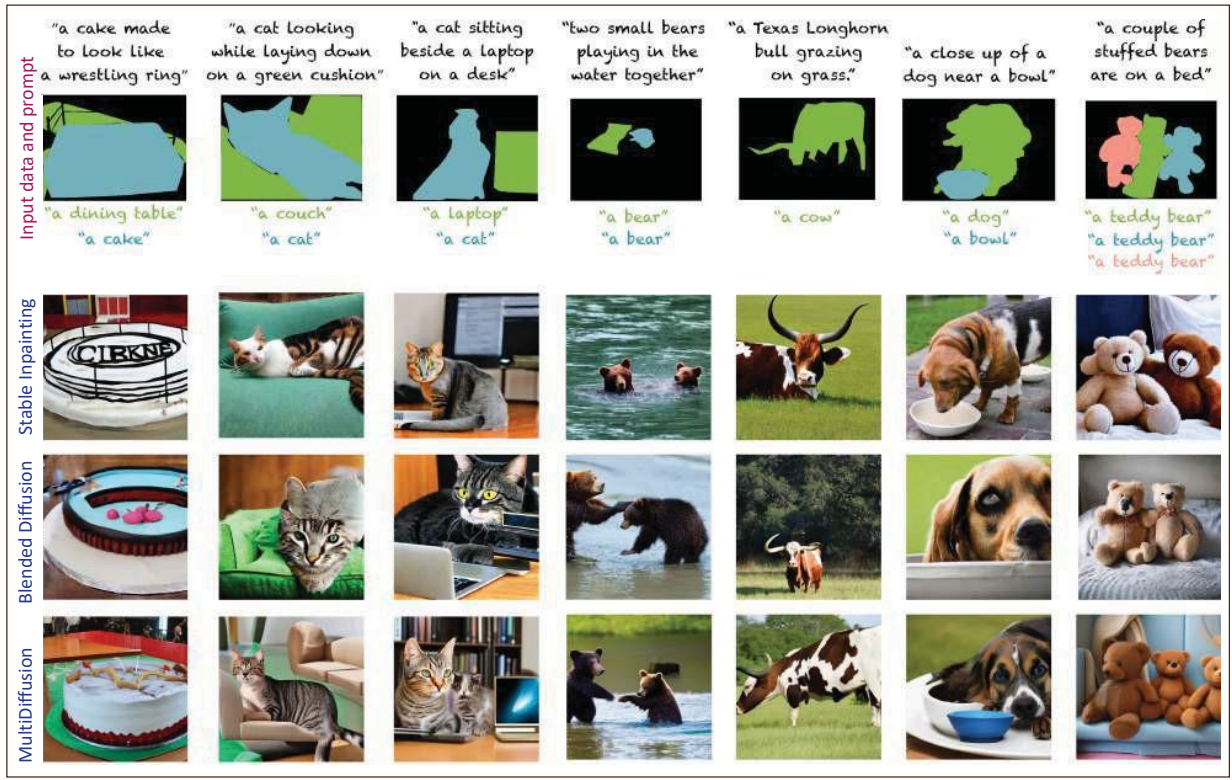

Figure 17. Examples of images generated by state-of-the-art diffusion models: Blended Diffusion [212], Multidiffusion [213] and Stable Inpainting [210]. The images have been generated using text prompts together with simple foreground images as inputs. Image is from Multidiffusion [213].

Table 3
Performance results of automated and traditional data augmentation methods on classification tasks. The values represent percentage accuracy on CIFAR-10, CIFAR-100 and ImageNet datasets. **Bold** is highest score, *italic* is second highest, and underline is third highest.

| Method | Classification accuracy on various datasets | | | |
|---|---|---|---|---|
| | CIFAR-10 | CIFAR-100 | ImageNet (top 1) | ImageNet (top 5) |
| Baseline | 96.1 | 81.2 | 76.3 | 92.1 |
| **Automated data augmentation methods** | | | | |
| TrivialAA [224] | 97.5 | <u>84.3</u> | 78.1 | 93.9 |
| DivAug [225] | *98.1* | 84.2 | 78.0 | - |
| AWS [153] | 98.0 | *84.7* | <u>79.4</u> | *94.5* |
| MetaAug [226] | 97.7 | 83.8 | 79.7 | **94.6** |
| PBA [159] | 97.4 | 83.3 | 77.2 | 93.4 |
| AdvAA [149] | *98.1* | 84.5 | **79.9** | *94.5* |
| A2-Aug [227] | 98.0 | **85.2** | 79.2 | - |
| FastAA [142] | 97.3 | 82.8 | 77.6 | 93.7 |
| KeepAA [228] | **98.7** | - | 78.0 | 93.0 |
| DeepAA [229] | 97.4 | 83.7 | 78.3 | - |
| **Traditional data augmentation methods** | | | | |
| StochasticDepth [230] | - | - | 77.5 | 93.7 |
| RE [231] | 96.9 | 82.3 | 77.3 | 93.3 |
| RICAP [232] | 97.2 | 82.0 | 78.6 | - |
| SaliencyMix [233] | 96.0 | 80.5 | 78.7 | - |
| GridMask [144] | 97.2 | 83.4 | 77.9 | - |
| SmoothMix [234] | - | - | 77.7 | 93.7 |
| Manifold Mixup [235] | | | 77.5 | 93.8 |
| CutOut [236] | 96.9 | 81.6 | 77.1 | - |
| MixUp [237] | 97.3 | 82.1 | 77.9 | 93.9 |
| FMix [238] | 96.4 | 82.0 | 77.7 | - |

# 5 Automated feature engineering

Feature engineering is a set of low-level data processing tasks that aim to increase the representativeness of input data by indirectly manipulating extracted feature vectors in deep learning layers. The goal is to obtain the minimum subset of features that can adequately encode all necessary information about the data without loss of predictive performance.

Traditional approaches to feature engineering [239] rely on assumptions about the underlying data and feature statistics to enrich feature representation. They typically employ analytical formulations of feature transformation operations to enhance the representativeness of extracted features. These techniques, apart from being enormously computationally expensive and heavily dependent on domain expertise, do not often lead to an optimal solution. For this reason, automated machine learning is considered as a more viable approach.

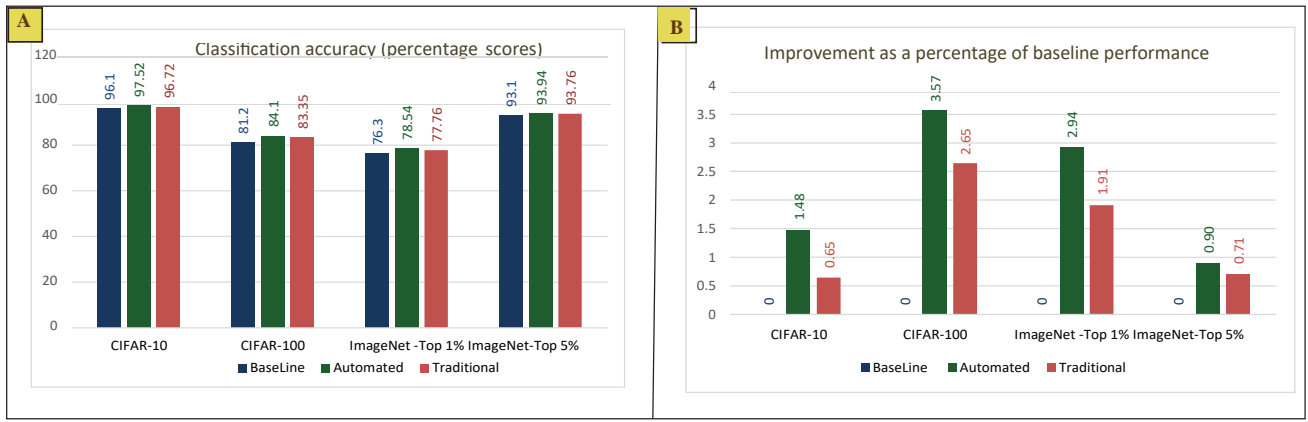

Figure 18. Comparison of automated data augmentation and traditional methods. Results for each class of approaches are based on 10 state-of-the-art techniques.

## 5.1 General concept of automated feature engineering

The basic idea of automated feature engineering is to create a large search space of all possible feature processing operations and then apply optimization techniques to search for an optimum feature engineering strategy. An optimum strategy is the best possible combination of the elementary operations (i.e., the set of operations that result in the best performance for the specific dataset and task).

Formally, for a target task $T_{sk}$, the feature engineering problem for a given dataset $D$ consisting of a set of features $F$ $f_1$, $f_2$, $f_3$, ...$f_n$ involves defining and applying a set of suitable transformations $T$ $t_1$, $t_2$, $t_3$, ...$t_n$ on the original feature set $F$ to generate additional feature set $\Phi$ and, with the help of a search mechanism, finding the best possible subset of features over the feature space $\Omega = F + \Phi$ that maximize performance on the target task $T_{sk}$.

Because of the need to run and evaluate an unusually large number of operations on the given data, and owing to complex interactions among these operations, it is often challenging to obtain a good solution in reasonable time, especially when dealing with large datasets. Instead of dealing with the entire search space at a go, it is often possible to create smaller search spaces which can be evaluated first before testing various combinations of the best performing strategies. A number of recent works, for example, [240], [241], have tended to favor such an approach. For example, VolcanoML, introduced by Li et al. in [240], aims to alleviate the challenge of computational burden of large search spaces by proposing a more scalable search space design that allows users to construct custom textitexecution plans for their specific AutoML tasks.

Instead of attempting to search on the entire search space for the optimum solution, the authors suggest decomposing it (the search space) into smaller components and then learn a strategy to select the most appropriate ones for a given task and dataset. To achieve this, they developed a high-level structural representation of the search space consisting of small atomic subspaces that can be used as building blocks for larger search spaces. The small search spaces form a tree structure and can be combined in different permutations. This provides a more flexible mechanism that allows selecting a final search space configuration based on computational budget and training time constraints. The authors demonstrated the ability of the approach to compose more efficient AutoML pipelines than with traditional search space configurations. NAS-based feature engineering approaches [59], [60] differ from conventional AutoML methods in the sense that they do not only optimize parameters and hyperparameters for predefined model architectures but are also concerned with creating, training and validating new neural network topologies based on the input data. The goal is to find fundamentally new model structures which perform better than manually designed architectures. In this case, for the construction of the search space, dynamic feature processing operations are defined together with elementary building blocks for generating surrogate neural architectures. Search techniques are the applied to jointly optimize the resulting neural architecture and the feature processing operations.

While a wide variety of AutoML tools (e.g., [97], [242], [243]) exist for generic data processing, dedicated AutoML-based feature engineering utilities and software packages [244], [245], [246], [247] have also been introduced that perform automated feature extraction as their core functions. A large number of these tools are aimed at processing features from time series and tabular data. Tools for time series data include TSFEL [248], tsfresh [246], tsfeaturex [247]. Featuretools [244] is aimed at extracting useful features from hierarchical and relational databases. This is achieved using a so-called deep feature synthesis that mines and combines features from multiple database tables at different hierarchical levels.

## 5.2 Common feature engineering tasks and approaches for automation

In the literature, three main feature engineering tasks can be distinguished [249]: feature extraction, feature synthesis

(often referred to as feature construction in many literature sources (e.g., [250], [251], [252], [253]), and feature selection. There is no consistent definition of these terms. Also, the scope and interrelationships among these three processes often vary widely. While many authors treat all three tasks as distinct processes, some authors (e.g., [254]) consider feature synthesis and selection to be aspects of the same process - feature extraction. And while authors such as Horn et al. [245] treat the extraction and synthesis subtasks as the same task, other literature sources (e.g., [255]) regard feature construction and extraction as a single process distinct from feature selection. In this work, we use the following designation:

- Feature extraction - relates to the process of manipulating input data or intermediate feature sets to obtain more generalizable, robust, representative and compact features for the particular task.
- Feature synthesis – also commonly termed feature construction [254], is the creation of new feature sets based on the available data or intermediate features to better characterize the dataset and enhance the performance of the resulting model.
- Feature selection – is the process of choosing and further processing only the best set of features that result in noticeable performance improvement.

Despite this clear delineation, the boundaries between these processes can be very blur, and in some cases, the terms can be used interchangeably. Moreover, they are often applied simultaneously in a given machine learning problem.

### 5.2.1  Automated feature extraction

For many large-scale machine learning problems, feature extraction is the first and most important stage in the feature engineering process. It allows to obtain more compact and representative feature set from raw data. Since the input data in these ML application settings are usually characterized by a large number of variables, it is generally useful to simplify feature representation by maintaining only the important variables that account for performance. The process of feature extraction results in new features with reduced computational and memory footprint but with superior characteristics necessary to encode the underlying data. Feature extraction can also enhance interpretability since the reduced set generally contains only significant features useful for the given task. The extracted features are generally obtained by applying various linear combinations of the original ones. This allows the dimensionality of identified set of features to be reduced while maintaining the information encoded by the original dataset. Traditional feature extraction techniques rely on algorithms such as principal component analysis (PCA) [256], [257], independent component analysis (ICA) [258], [259], linear discriminant analysis (LDA) [260] and locally linear embedding (LLE) [261], [262]. In AutoML pipelines, feature extraction is usually achieved by using simplified analytical functions based on some variations of these standard algorithms to perform basic feature extraction operations (e.g., in [263], [264]) within a bi-level optimization scheme, where the best performing operations and the corresponding hyperparameter values are automatically determined. Because of the potential for excessive explosion of features, most of these approaches fuse the extraction and selection stages as a single process. Yang et al. [265], for instance, employ simple mathematical operations to first extract data and then utilize Boruta algorithm proposed in [264] to select the most effective ones.

Some new approaches [266], [267], [268], instead of focusing solely on utilizing analytical functions to extract features from data, propose to rather learn effective network architectures for applying the necessary feature extraction operations. These approaches are based on the concept of neural architecture search (NAS). For instance, Meta-learning for Tabular Data (METABU) [267] employs a NAS-based technique to automatically extrct useful features by finding the best machine learning model and the corresponding hyperparameter values. Firstly, the authors manually constructed a large number of meta-features (135 in all). They then employ optimal transport (OT) [269] technique to perform linear transformation on these manually designed meta-features in order to generate new features that are linear combinations of the basic meta-features. Finally, AutoML technique is used to find the best model configuration and hyperparameter values that results in the best features based on performance as measured on a validation set. Also in [266], Lopes et al. propose an automatic feature extraction method for sentiment analysis problem based on Neural Architecture Search (see Figure 19). The authors constructed diverse machine learning models, each of which independently performs feature extraction in different ways. They then used random search techniques to find the best performing model based on the performance scores on validation data. The approach uses a three-stage structure to accomplish the task of predicting sentiments based on enhanced features extracted from input data. The model takes as input images obtained from the internet and their corresponding textual descriptions. After an initial preprocessing stage, image and text data are classified separately by predefined models in the search space. The image and textual features are then fused together to obtained a composite feature representation before categorizing the underlying inputs into one of the following: Positive, Neutral and Negative. The fusion process, in this formulation, is essentially a feature extraction mechanism that produces useful features from combined text and image data.

Transferable AutoMl (Tr-AutoML) [268] utilizes a meta-feature extraction method that allows useful features learned from previously trained, multiple AutoML pipelines to be mined, aggregated and then transferred to new tasks. The approach combines meta-learning and architecture search for feature extraction from multiple but related datasets. It relies on transferring already-learned architectures that previously showed good results in performing feature extraction functions on new datasets and tasks. This significantly reduces the search space since a common architecture can serve for multiple feature extraction tasks.

### 5.2.2  Automated feature synthesis

The feature synthesis or construction stage creates additional features to further enhance the performance of the resulting model. The synthesis process generally exploits the statistical distribution of features extracted from the data to generate new, complementary features. This is useful in cases where extracted features are insufficient to provide an adequate representation of the input data. Typically, feature synthesis is achieved by performing manipulations on accessible features with the help of predefined transformation operations. Apart from generating new features, feature synthesis models [40], [253], [271] typically incorporate mechanisms to perform feature selection as not all artificially constructed features would be useful for the given task. However, these approaches, as discussed in this subsection, differ from feature selection methods as their main focus is in generating good features as oppose to merely selecting useful features from existing set extracted directly from input data.

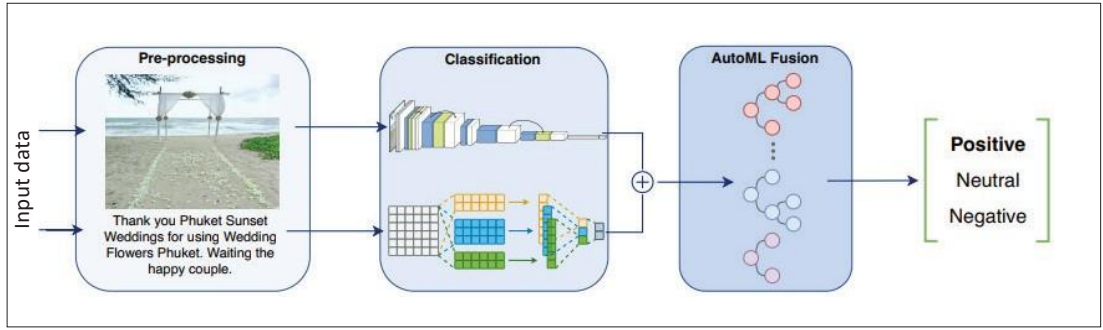

Figure 19. Functional diagram of the automated feature extraction techniques proposed in [266]. The approach employs a three-stage neural network framework to perform sentiment analysis by first extracting and separating online images and their corresponding labels. Different models are then used to perform preprocessing before sentiment classification, also by independent sub-models. Different combinations of these models are then evaluated and the best settings selected according to the performance on the target task.

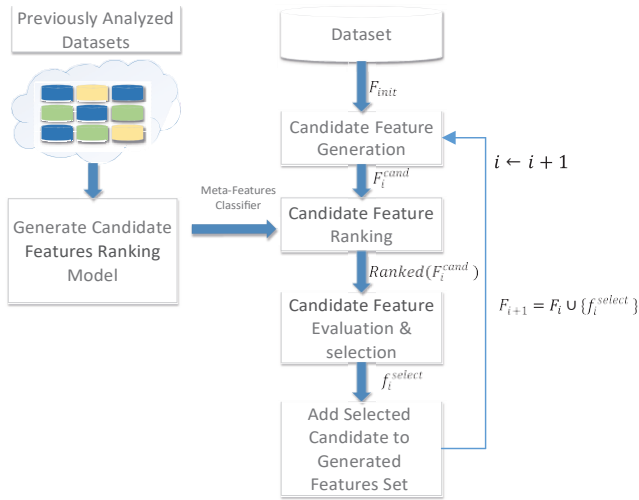

Figure 20. Basic principle of feature synthesis and selection approach in Explorekit [270]. This relies on drastically increasing the feature set by generating additional features and then searching for a small subset of features the yield the best performance.

Common techniques used to accomplish this task include feature transformation, interpolation, averaging and mixing. Traditional approaches to solving this problem requires repetitive trial and error using different combinations of operations with different hyperparameter settings. Automated feature synthesis [38], [245], [272], [273], [274] is particularly valuable in situations where the scarcity of data restricts the adequacy of features that are mined through feature extraction process. For a given feature set derived from a training data, the automated feature synthesis task seeks to determine and optimize the best combination of basic feature-level processing operations and their associated hyperparameters to apply in order to generate new features that, when incorporated into the original feature set, maximize the performance of the model on the target data. Some automated feature synthesis methods (e.g., [274], [275], [276]) apply all operations at once to expand feature set and then search for the best features within the expanded set. Kanter and Veeramachaneni [274], for instance, simultaneously apply all transformation operations in the search space and then utilize different classification sub-models to select the most effective features. The main drawback of this approach is the high memory requirements that results from the need to store a large pool of features.

Another line of works (e.g., [270], [273]), instead of aggressively expanding the search space, iteratively apply small number of operations to expand, test and select small subsets of features in batches. While this approach overcomes the high memory demand of techniques such as [274], [275], [276], testing for effective features at each iteration can lead to expensive computational overheads. Moreover, useful features that could form the basis to further generate more effective features can be lost in the selection process if they do not directly yield good performance. Thus, techniques based on this concept can fail to achieve optimum feature,,. In order to reduce the high computational cost of evaluating all candidate features, more recent approaches [271], [277] suggest using meta-learning techniques to predict effective transformations that generate the best features and to prioritize them for the synthesis and evaluation stages.

Different from many of the feature synthesis paradigms [271], [277] that employ a bi-level optimization mechanism to generate useful features, some approaches [270], [276], [278] utilized more conventional deep neural networks to accomplish feature engineering. These models learn to predict the effectiveness of different transformation operations through previous training on other datasets. Some works such as [59] have demonstrated the effectiveness of NAS techniques to synthesize useful features. These works construct the search space consisting of basic neural models and then employ neural architecture search approaches to find the best models capable of applying feature transformations to generate new features.

### 5.2.3 Automated feature selection

The overall goal of feature selection is to find a minimal set of features that provides the best predictive performance using the least possible computational resources. The process, thus, effectively simplifies the resulting model by eliminating irrelevant features. The overall performance can be drastically enhanced since irrelevant features often harms performance. In addition, feature selection helps to determine the effectiveness of different feature sets on the ML model and hence can provide additional insights about the important factors that influence performance [57]. Moreover, the reduction of features by selecting only the most useful ones leads to a more simplified and interpretable models. Feature selection, thus, performs four important functions:

- Model size reduction by the elimination of redundant and less useful features, leading to a reduction in computational and memory demand
- Predictive performance improvement as a result of the elimination of potentially detrimental features
- Model structure simplification as a result of reduced feature set
- Knowledge discovery (i.e., provides insights into the internal representation)
- Interpretability of data and resulting machine learning model

A major difficulty with feature selection is the fact that it is often not straightforward to determine the minimum subset of features that can provide optimum performance. A set of features that may not be useful when applied alone may turn out to be the optimal set when used in combination with another set of features. This makes the feature selection process an inherently combinatorial problem. Moreover, there may be multiple sets of features that result in optimal performance. Feature selection approaches aimed at knowledge discovery [57], [279], [280], [281], [282] often need to identify all such minimal feature sets. However, in some scenarios it may be necessary to identify all such sets of features. For example, feature selection for knowledge discovery applications generally aims to identify all useful sets of features. On the other hand, for problems such as regression, classification and time series forecasting, it is important to be able to quickly find only one out of the most useful sets of minimal features that can comprehensively encode the given data.

Typically, to carry out feature selection, a subset of features is drawn from the set of learned features based on a given criterion and then evaluated to assess the efficacy of the chosen subset on the end task. The process is repeated until a small subset of features is obtained that can provide optimal performance. For automated feature selection, like in most automated machine learning tasks, optimization methods are usually employed to find the optimum feature sets. Explorekit [270] (20) uses a learned mechanism that employs a ranking classifier to identify useful features for further exploitation. In conventional deep learning models, regularization techniques based on feature dropout methods [283] are a common way to eliminate irrelevant features and retain more useful ones. But hyperparameter selection for these dropout mechanisms is generally accomplished manually, making it challenging to find the optimum set of features to drop in order to retain only the most effective set. Inspired by these feature-level information dropout regularization techniques, some recent feature engineering methods [283], [284] are aimed at learning optimum dropout patterns based on automated hyperparameter tuning techniques. These approaches can be considered as an alternative variant of automated feature selection methods since, by eliminating irrelevant features, they practically allow only the most useful subset of features to be selected for the given task. AutoDropout [285], proposed by Pham and Le, learns effective feature-level information dropout using AutoML-based hyperparameter tuning. To accomplish this, the authors created primitive dropout patterns using adjustable hyperparameters that define important properties of the dropout pattern, including the dropout size, stride, and the number of times to repeat the particular pattern. In addition, hyperparameters describing geometric operations, viz. rotations and shearing, as well as additional flags (binary variables) that determine whether or not dropout patterns are shared across channels or are applied on the residual branches of ResNet, are also specified. In the optimization stage, optimum settings for variables describing these hyperparameters are then automatically learned through feedback of the loss signal.

## 5.3 End-to-end automated feature engineering

While individual feature processing tasks can be automating separately, many modern automated feature processing approaches [134], [274], [275], [286], [287], [288] are aimed at performing the full range of feature engineering tasks (i.e., feature extraction, synthesis of additional features, and selection of useful features) in a holistic and end-to-end manner. In this works, the performance of the various subtasks may not be distinctive. Most approaches generally follow the so-called expansion-reduction paradigm, where various transformations are applied to the extracted features to create additional features that are then evaluated together with the original features and the best set of features selected. These idea is to utilize a composition of primitive transformation operations within a machine learning pipeline to generate diverse features which are jointly optimized with model hyperparameters in the training process using various search mechanisms. For instance, Scalable Automatic Feature Engineering (SAFE) [40] (Figure 21) employs a search strategy based on XBoost [289] to find the best combinations of useful features. The search is conditioned on maximizing information gain. Evolutionary Automated Feature Engineering (EAAFE) [286] and Ring Theory based Harmony Search (RTHS) [290] employ evolutionary computational algorithms to find useful features. The use of reinforcement learning approaches is also common [276].

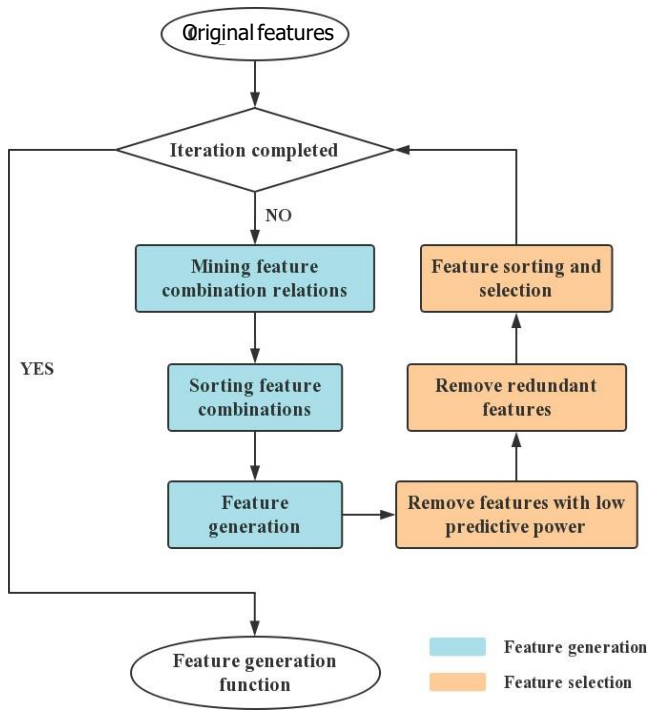

Figure 21. Scalable Automatic Feature Engineering (SAFE) [40] exploits the relationships among the initial feature set to obtain a reduced feature set which is then expanded and further filtered according to the performance of the various combinations of features generated.

Neural feature search (NFS) [59] uses a reinforcement learning-based RNN controllers to learn effective transformation policies for generating useful features. Differentiable Automated Feature Engineering (DIFER) [288] formulates the automated feature engineering task as a differentiable problem by mapping discrete features into a continuous vector space with the help of a so-called *encoder-predictor-decoder-based feature optimizer* and then finds effective features by gradient ascent. The general architecture of their framework is shown in Figure 22. The approach uses random sampling to generate initial population of features which are further transformed by applying a sequence of transformation operations.

### 5.4 Automated feature engineering based on NAS frameworks

Automated feature engineering methods [59], [59], [291] have also been proposed that employ Neural Architecture Search (NAS) techniques for feature selection. For instance, Chen et al. in [291] employ a NAS mechanism to search for and select the most effective set of features for a multimodal person re-identification system based on RGB and infrared images. Their technique incorporates a two-level search space construction, where spatial and channel-wise features are independently selected. Neural Feature Search (NFS) [59] proposes a method that utilizes NAS technique to accomplish feature engineering (actually extraction and selection). The authors utilized reinforcement learning algorithm to train a Recurrent Neural Network-based controller to generate useful transformations. The technique first aims to find a neural network model capable of performing the most effective feature transformations. The resulting features are selected and then enhanced by applying subsequent feature engineering operations. This process alleviates the problem of feature explosion where redundant and less useful features are spawn that adversely affect predictive performance while at the same time contributing to computational complexity and memory demands. Similarly, Rakotoarison et al. in [267] propose to learn the best values for model hyperparameters in order to optimize the selection of manually composed meta-features. The authors show that meta-features learned using AutoML techniques can outperform state-of-the-art feature extraction methods based on traditional machine learning techniques.

### 5.5 Performance of automated feature engineering methods

In this section, we compare performance results of automated and traditional feature engineering techniques commonly employed in practice and in the scientific literature. The first set of results (shown in Table 5) show classification accuracies (F-1 scores [292]) obtained from experimental evaluations by several authors –specifically, [59], [278], [288], [293]. The feature engineering methods compared are the following. *Raw* – this entails using the original feature set without any further feature processing. Random (*Ran*) involves iteratively running through the feature set and applying a random set of transformation operations (or no transformation) on the features in each iteration. In the case of *ME*, the feature engineering algorithm selects only one transformation operation – the most effective operation – for each sample set. Brute-force (*BF*) applies all transformation operations on the entire feature set and then finds the best subset of features from the expanded set. In the literature, this approach is commonly referred to as *Expansion-Reduction*. The rest of the methods are LFE [278], NFS [59], AutoFeat [245] and DiFFER [288]. We selected 23 datasets where performance results are widely available for many of the aforementioned automamted feature engineering approaches. These datasets are from UCI [294], [295] and OpenML [296] repositories. For more detailed information about their basic characteristics, the reader may refer to their repositories [294] and Github pages [295], [297]. Random Forest model is employed as the base model for the classification tasks using ten-fold cross validation.

From the table, it can be seen that the automated feature engineering techniques clearly outperform basic methods based on traditional paradigms. However, it is worth noting that the high performance of automated methods is not consistent for all datasets. Particularly, *spambase*, *autos* and *convex* show little or no improvements with automated feature engineering methods.

We present a composite table (Table 6) about the experimental performance evaluations in [293]. The results show the comparative performance of state-of-the-art traditional feature engineering methods (DCN-V2 [298] and FCTree [299]) and automated engineering methods (FETCH [300], AutoCross [301], SAFE [40], OpenFE [293] and AutoFeat [245]). Baseline results (i.e., task performance without fea-ture engineering) are included for comparison. The evaluated models are implemented using LightGBM [302]. Details of the datasets are presented in [293], [303], [304]. The models are run ten times on each of the datasets. It can be observed that in most of the cases the traditional approaches fail to outperform the baseline. While traditional approaches have shown strong performance over baseline results on UC Irvine and OpenML datasets in several works, including [59], [278], [288], [293], they appear to perform very poor per [293] (shown in Table 6). This is due to the fact that feature engineering process is highly dataset-sensitive, and the approaches in DCN-V2 [298] and FCTree [299] were optimized for datasets, which are fundamentally different from the UC Irvine and OpenML datasets. Indeed, for a number of the dataset in Table 6, the performance of the automated methods also dropped significantly, sometimes below the baseline. Li et al. [300] acknowledged this problem, and their approach, Feature Set Data-Driven Search (FETCH), aims to specifically tackle this issue by developing a more generalizeable feature generation and selection mechanism.

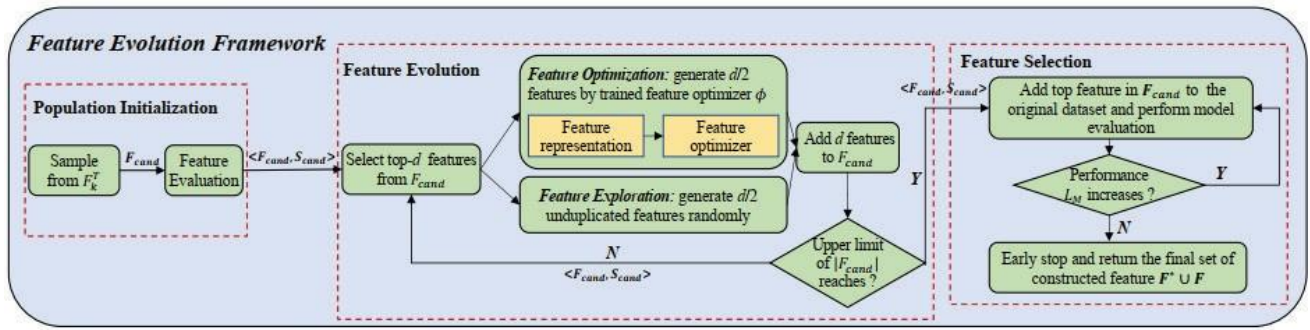

Figure 22. The basic design of differentiable Automated Feature Engineering (DIFER) [288] is made up of three main components: (1) population initialization module which generates feature by sampling randomly from the initial set, (2) feature evolution module for growing and introducing diversity into the feature set, and (3) feature selection which aims to reduce the overall feature count by selecting only the best set of features.

Table 4
Degrees of automation of feature engineering tasks.

| Method | Description and main characteristics | Example works |
|---|---|---|
| Analytical | Basic and rigid automation based on fixed analytical formulas to separately extract, synthesize and select useful features | [260], [261], [262] |
| DL -based | Deep learning-based approach to generate and select useful features given a set of raw data in an end-to-end manner. | [270], [276], [278] |
| Fully automated | AutoML and NAS-based approaches to holistically generate useful features directly from raw data, together with model selection and hyperparameter optimization. | [274], [286], [287] |

Table 5
Performance of state-of-the-art feature engineering methods. The results are classification accuracies (F-1 scores) on selected datasets from the UCI [294], [295] and OpenML [296] repositories. **Bold** is highest score, *italic* is second highest, and underline is third highest. NB: Feat. -number of features, Inst. - number of samples.

| Dataset | Feat. | Inst. | Raw | ME | BF | Ran | NFS | Auto-Feat | LFE | Differ |
|---|---|---|---|---|---|---|---|---|---|---|
| AP-0mentum-lung | 10936 | 203 | 0.883 | 0.915 | underline{0.925} | 0.908 | **0.981** | - | *0.929* | |
| AP-0mentum-ovary | 10936 | 275 | 0.724 | 0.775 | 0.801 | 0.745 | **0.873** | - | underline{0.811} | *0.8724* |
| autos | 48 | 4562 | *0.946* | 0.95 | underline{0.944} | 0.929 | - | - | **0.96** | |
| Balance-scale | 8 | 369 | 0.884 | *0.916* | underline{0.892} | 0.881 | - | - | **0.919** | - |
| convex | 784 | 50000 | *0.82* | 0.5 | **0.913** | 0.5 | - | - | underline{0.819} | |
| Credit-a | 6 | 690 | 0.753 | 0.647 | 0.521 | 0.643 | underline{0.803} | *0.8391* | 0.771 | **0.8826** |
| dbworld-bodies | 2 | 100 | 0.93 | 0.939 | underline{0.927} | 0.909 | - | - | **0.961** | - |
| diabetes | 8 | 768 | underline{0.745} | 0.694 | 0.737 | 0.719 | **0.786** | - | 0.762 | |
| fertility | 9 | 100 | 0.854 | 0.872 | 0.861 | 0.832 | **0.913** | 0.7900 | underline{0.873} | *0.9098* |
| gisette | 5000 | 2100 | 0.941 | 0.601 | 0.741 | 0.855 | *0.959* | - | underline{0.942} | **0.9635** |
| hepatitis | 6 | 155 | 0.747 | 0.736 | 0.753 | 0.727 | **0.905** | 0.7677 | underline{0.807} | *0.8339* |
| higgs-boson-subset | 28 | 5000 | underline{0.676} | 0.584 | 0.661 | 0.663 | **0.827** | - | 0.827 | |
| ionospare | 34 | 351 | 0.931 | 0.918 | 0.912 | 0.907 | *0.972* | 0.9117 | underline{0.932} | **0.9770** |
| labor | 8 | 57 | underline{0.856} | 0.827 | 0.855 | 0.806 | **0.960** | - | *0.896* | - |
| lymph | 10936 | 138 | underline{0.673} | 0.664 | 0.534 | 0.666 | **0.987** | - | 0.757 | - |
| madelon | 500 | 780 | underline{0.612} | 0.549 | 0.585 | 0.551 | **0.836** | - | 0.617 | - |
| Megawatt1 | 37 | 253 | 0.873 | 0.874 | 0.882 | 0.869 | **0.933** | underline{0.8893} | 0.894 | *0.9171* |
| Pima-indians | 8 | 768 | 0.74 | 0.687 | 0.751 | 0.726 | **0.790** | underline{0.7631} | 0.745 | *0.7865* |
| Secom | 590 | 470 | 0.917 | 0.917 | 0.913 | 0.915 | **0.934** | - | 0.914 | - |
| sonar | 60 | 208 | 0.808 | 0.763 | 0.468 | 0.462 | **0.839** | - | underline{0.801} | - |
| spambase | 57 | 4601 | **0.948** | 0.737 | 0.39 | 0.413 | **0.948** | - | underline{0.947} | - |
| Spectf-heart | 43 | 80 | 0.941 | **0.955** | 0.881 | *0.942* | - | - | **0.955** | - |
| Twitter-absolute | 77 | 140707 | **0.964** | 0.866 | 0.946 | *0.958* | - | - | **0.964** | - |

Table 6
Performance of automated feature engineering methods (FETCH [300], AutoCross [301], SAFE [40], OpenFE [293] and AutoFeat [245]) against traditional and baseline methods methods (DCN-V2 [298] and FCTree [299]). **Bold** is highest score, *italic* is second highest, and <u>underline</u> is third highest. NB: No.S (k) - number of samples (in 1000s); CL - number of classes; Cat, Ord and Num - numbers of categorical, ordinal and numerical features.

| Dataset | No.S (K) | Cl | Features | | | Metric | Raw | DCN-V2 | FC-Tree | SAFE | Auto-Feat | Auto-Cross | Open-Fe | FET-CH |
|---|---|---|---|---|---|---|---|---|---|---|---|---|---|---|
| | | | Cat | Ord | Num | | | | | | | | | |
| Vehicle | 98.5 | 2 | 0 | 0 | 100 | AUC^ | 0.925 | 0.924 | <u>0.926</u> | 0.925 | 0.925 | 0.921 | **0.928** | *0.927* |
| Diabetes | 102 | 2 | 34 | 10 | 3 | AUC^ | 0.731 | 0.717 | 0.731 | 0.730 | 0.732 | 0.732 | **0.888** | 0.731 |
| Telecom | 51 | 2 | 22 | 14 | 21 | AUC^ | 0.671 | 0.661 | 0.671 | 0.673 | 0.672 | 0.651 | **0.680** | *0.673* |
| California Housing | 20.6 | - | 0 | 1 | 7 | RMSE_ | <u>0.432</u> | 0.479 | 0.432 | - | 0.444 | - | **0.421** | *0.430* |
| Microsoft | 1200 | - | 0 | 25 | 111 | RMSE_ | *0.744* | 0.750 | 0.744 | - | <u>0.744</u> | - | **0.738** | X |
| Nomao | 34.4 | 2 | 29 | 55 | 34 | AUC^ | *0.996* | 0.992 | *0.996* | 0.996 | 0.996 | 0.993 | **0.997** | *0.996* |
| Broken Machine | 900 | 2 | 0 | 27 | 31 | AUC^ | <u>0.756</u> | 0.748 | 0.750 | 0.750 | 0.750 | 0.765 | **0.786** | X |
| Jannis | 83.7 | 4 | 0 | 0 | 54 | Acc.^ | *0.721* | 0.720 | 0.719 | - | *0.721* | - | **0.729** | 0.720 |
| Covertype | 581 | 7 | 0 | 45 | 9 | Acc.^ | *0.969* | 0.966 | 0.719 | - | <u>0.968</u> | - | **0.974** | X |
| Medical | 163 | - | 6 | 0 | 5 | RMSE_ | <u>1128</u> | 1413.7 | *1089* | - | 1172 | - | **982.0** | 1130.4 |

# 6 HOLISTIC, END-TO-END WORKFLOW OF DATA PROCESSING IN MACHINE LEARNING

In Sections 3 through 5, we describe approaches for automating specific low-level data processing tasks such as data cleaning, labeling and feature engineering. Many automated machine learning techniques incorporate manual processing in some stages of the pipeline. For example, feature engineering may be fully automated but appropriate transformations can be manually selected to pre-process input data according to the specific details of the task and the properties of the underlying data (e.g., in [40], [286], [288]). Recently, there is a trend toward full automation of the entire machine learning pipeline from data acquisition to model deployment, where not only data-centric tasks are automated, but also the realization of the overall big data solution (Figure 23).

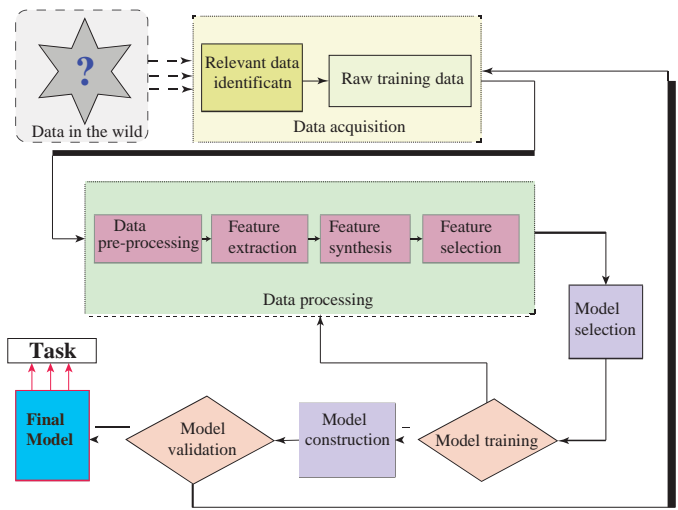

Figure 23. Simplified architechture of an end-to-end AutoML framework.

End-to-end frameworks aim to automatically collect and process data, perform feature engineering and also fix various errors inherent in the data (for example, missing metadata) or errors that may arise in the course of processing. In addition, these frameworks can simultaneously carry out hyperparameters optimization, model architecture construc-tion, selecting of evaluation metrics, prediction, analysis of results, and performing other machine learning tasks. This enhances the ability to discover valuable semantic information from large volumes of unstructured data for big data applications.

To enable a single end-to-end AutoML framework to solve a wide range of big data problems, several components are often used to create AutoML systems. In particular, different blocks of processing units are typically used to handle different tasks such as data cleaning, augmentation and feature engineering. These functional blocks are often arranged sequentially in the form of a structured pipeline. To deal with the different processing requirements of diverse data formats and tasks, multiple pipelines with different structures incorporating several machine learning models are often used.

Since the types of computational blocks required for a particular problem depends on the nature of dataset and task, the selection of the structure and parameters of the specific pipeline to use is an important part of the automated data processing and machine learning tasks. Thus, the automation problem reduces to the task of combining and optimizing the various models in the pipeline in order to achieve the best possible performance. The pipeline is often represented as a computational graph, and its structure is determined using various functions to numerically evaluate the performance of different configurations. While some end-to-end AutoML systems such as ATM [305], ML-Plan [306] and Hyperopt-Sklearn [307] utilize fixed pipelines, a large number of new approaches (e.g., AutoDES [308], FLAML [309], RECIPE [310] and H2O AutoML [100]) employ variable-length pipelines. Approaches employing variable pipelines typically use evolutionary computational algorithms to create a population of basic processing operations and iteratively adapt the simple algorithms to generate new and better algorithms through mutation and crossover. Recently, many large-scale generic AutoML tools have been developed to allow off-the-shelf application of machine learning models for various data-centric tasks. The main categories, characteristics and functions of these systems are presented in Section 8.

AutoSmart [312] (Figure 25) is another example of a fully automated machine learning framework that performs several data-centric tasks in an end-to-end manner. These include data preprocessing, data integration (i.e., table merging), feature synthesis and selection, as well as ensemble learning and model hyperparameter tuning. The framework incorporates a computational time and memory controller for managing the usage of computational resources required for the task. There are many off-the-shelf tools built on the basis of end-to-end automated machine learning that are even more complex and functional more capable than the models described above. We describe these models in more detail in Section 8.

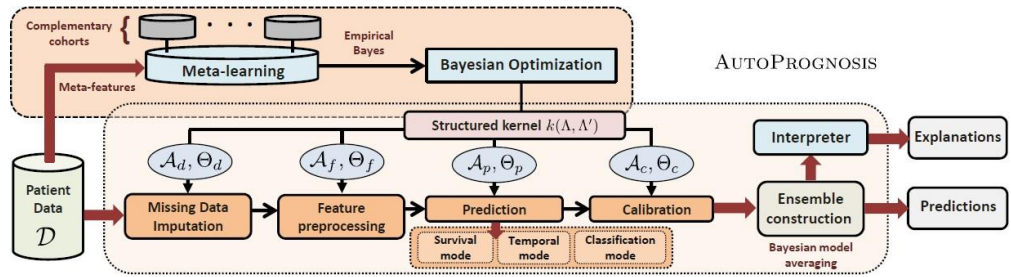

Figure 24. AutoPrognosis [311] uses patient data to perform end-to-end automated data processing. The system is aimed at enhancing clinicians' decision making by making accurate prognosis of health outcomes and also providing explanations for predictions.

## 7   GENERIC AUTOML TOOLS FOR DATA PROCESSING AND FEATURE ENGINEERING

Generic AutoML tools are aimed at enabling a fully automated workflow in model development. That is, allow non-professional users to use AutoML system to take raw input data, perform all necessary processing functions and then generate suitable models for an end application. Thus, they provide non-expert users the opportunity to use state-of-the-art machine level techniques to solve complex problems without explicitly processing data and building models. These tools are generally characterized by a high degree of flexibility. They allow to construct ML pipelines using various types of input data. The most commonly supported input formats are tabular, text, time series and image data types. Many of the tools offer a way to customize and manage model performance and complexity. Some of the most widely used generic AutoML frameworks include TPOT [96], Auto-WEKA [243], Amazon SageMaker Autopilot [313], DataRobot [314], AutoKeras [97], AutoGluon [315] and H2O AutoML [100]. We summarize the main features of these frameworks in Table 7.

While most of the most popular tools are designed to tackle general problems related to Big Data, some AutoML tools–e.g., Google AutoML Vision [242], AutoPrognosis [311], JADBio [57], GAMA [316] and MedicMind [27], Microsoft Azure Custom Vision [317]–are domain-specific. AutoPrognosis [311] and MedicMind [318], for instance, are specifically designed to deal with healthcare related data. AutoPrognosis [311] (shown in Figure 24) is concerned with using electronic health records to aid diagnosis while MedicMind [318] deals with medical image analysis. Google AutoML Vision [242] and Microsoft Azure Custom Vision [317] are aimed at solving computer vision problems such as image classification and object recognition.

### 7.1   Common functions supported by AutoML tools

In the course of training, the autoML framework creates several alternative pipelines that test, evaluate and validate different sets of machine learning algorithms and hyperparameter settings for solving specific problems. To accomplish this, the frameworks are usually sub-divided into specialized components dealing with specific  tasks and data types: image, text or tabular data.  There may also be special modules to support feature generation and preprocessing. AutoM tools typically handle multiple data processing tasks, including data collection and preliminary preparation, processing and post-processing: They can automatically find missing data, identify incorrect labels, and select the desired subsets of data required for a given  task.

The tools generally come with user-friendly graphic user interfaces (GUIs) with production-ready tools. The main functions include task analysis to understand the needs of the particular user; problem recognition – to discover the specific problem the user intends to solve and generate a set of potential machine learning tasks capable of solving the problem; model evaluation and validation are carried out to arrive at an optimal solution. The GUI and interface allow users to easily accomplish all these steps by performing intuitive actions to achieve desired results in an almost automated way; given a particular dataset, with minimal user input, the system is able to specify plausible problems that can be exploited to automatically generate models that produce end solutions. AutoML systems also provide intuitive information about the generated models and the data manipulation processes in a way that provides additional insights to guide non-expert users. Based on this information and their specific goals and preferences, users are able to refine the automatically defined problems and solution sets. Many AutoML platforms are readily compatible and interoperable with standard tools, allowing the results of the processing stage to be seamlessly integrated with external applications [18]. In addition to data processing functions, commercial AutoML tools provide additional functionalities such as data visualization. Tools like Pecan AI [332] allow users to seamlessly interact with their models and data without any coding, whatsoever.

Another important function commonly supported by advanced AutoML tools is visualization. Visualization utilities typically provide all necessary information needed to understand and make useful decisions about the task in relation to the available data. Such capabilities facilitate interpretability and provides useful insights into complex machine learning mechanisms. This is useful for knowledge discovery applications. In medicine and healthcare, these applications include the discovery of novel drug targets, or the design ofbetter assays with minimal field measuring and testing requirements. Visualization capabilities can also allow developers to better assess the robustness and overall quality of a model before deploying in practical applications. For instance, one can examine the effect of missing specific features by deliberately removing features of interest from a given model (e.g., in [57]). Moreover, visualization of the AutoML process can help to enrich the knowledge of data scientists or machine learning researchers by providing new insights [58]. It also enhances trust in AutoML solutions [333], [334] since developers can visually probe and validate pertinent issues related to trust and reliability in an intuitive manner.

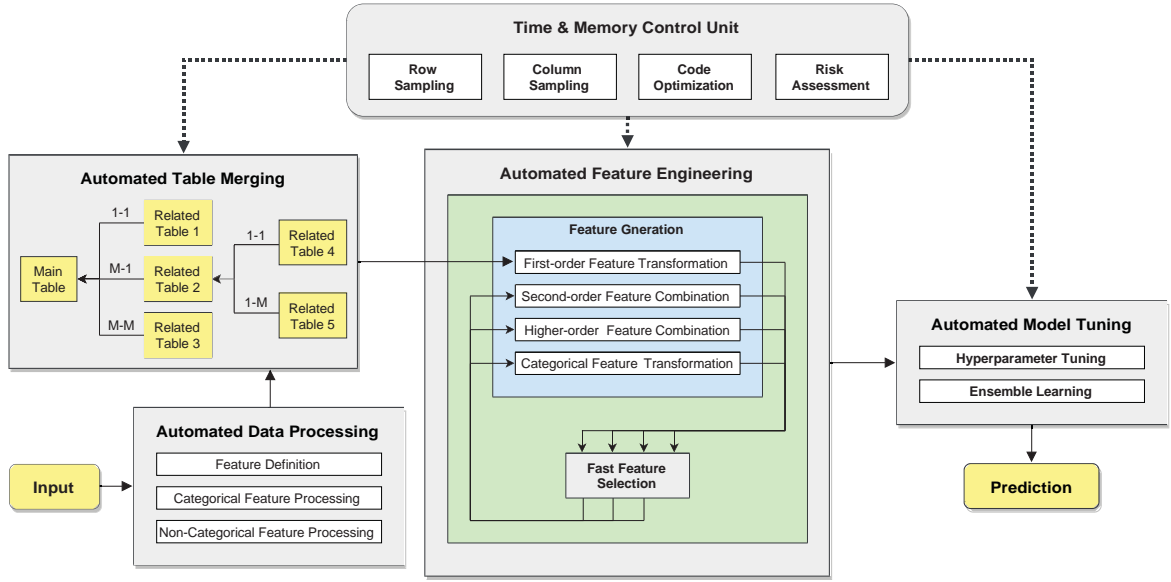

Figure 25. The general architecture of AutoSmart [312]. The approach performs several data preprocessing and feature engineering tasks in an end-to-end manner.

Table 7
A summary of the main features of popular generic AutoML tools indicating the main tasks they perform: CV-computer vision; NLP-natural language processing, Class. - classification, and Regres. -regression.

| Tool | Type | Data types supported | Main domains and tasks | | | | |
|---|---|---|---|---|---|---|---|
| | | | CV | NLP | Time series | Class. | Regr. |
| AutoKeras [97] | Open source | text, images, tabular | √ | √ | √ | √ | √ |
| Amazon SageMaker Autopilot [313] | Proprietary | Tabular, time series, images, texts | × | × | √ | √ | √ |
| Google Vertex AI [319] | Proprietary | tabular, image, text, and video data | √ | √ | √ | √ | √ |
| Google Cloud AutoML [242] | Proprietary | tabular, image, text, and video data | √ | √ | √ | √ | √ |
| Auto-WEKA [243] | Open source | Tabular | × | × | × | √ | √ |
| BigML OptiML [320] | Open source | Tabular, images, text, time series | √ | √ | √ | √ | √ |
| PyCaret [321] | Open source | Tabular, text, time series | √ | √ | √ | √ | √ |
| DataBricks [322] | Proprietary | Tabular, text, time series | | X | | √ | √ |
| AutoGluon (Amazon) [315] | Proprietary | image, text, tabular data | √ | √ | √ | √ | √ |
| Auto-sklearn [323] | Open source | Tabular | × | × | × | √ | √ |
| MS Azure AutoML [324] | Proprietary | time series, images, text | √* | √ | √ | √ | √ |
| DataRobot [314] | Proprietary | time series, images, text | √ | √ | √ | √ | √ |
| H2O AutoML [100] | Proprietary | Tabular, Texts | √ | √ | | √ | √ |
| H2O Driverless AI [325] | Proprietary | Time series, tabular, image, text | √ | √ | √ | √ | √ |
| TPOT [96] | Open source | Tabular, text*, image* | √* | | X | √ | √ |
| Darwin [326] | Proprietary | Tabular | × | × | | × | √ |
| IBM Watson AutoAI [327] | Proprietary | Tabular, text, time series | × | × | √ | √ | √ |
| FEDOT [328] | Open source | Tabular, time series, images, text | √ | √ | √ | √ | √ |
| MLBox [329] | Open source | Tabular | × | × | × | √ | √ |
| NNI (Microsoft) [330] | Open source | Tabular, images | × | × | × | √ | √ |
| Ludwig [331] | Open source | Tabular, text, time series | √ | √ | √ | √ | √ |
| FLAML (Microsoft) [309] | Proprietary | Tabular, text, time series | × | × | √ | √ | √ |
| ATM [305] | Open source | Tabular, image, text | | | | √ | √ |

Since many large-scale AutoML tools integrate several technologies, and multiple data pipelines into a single platform, they significantly increase synergy and effectiveness of solutions. The use of highly integrated platforms also speeds up development time by eliminating additional stages of development, testing and deployment of big data solutions.

## 7.2   Main categories and features of AutoML tools

AutoML tools can broadly be divided into two categories: open source and proprietary tools.

### 7.2.1   Open source AutoML tools

Common open source systems include Auto-Weka [243], AutoKeras [97], PyCaret [321], TROT [96], FEDOT [328] and Auto-sklearn [323]. The development of open source AutoML tools is characterized by the involvement of very large community of developers who contribute and update different functionalities independently. They are generally built on open source machine learning libraries and frameworks such us PyTorch [335], Scikit-learn [336]and TensorFlow [337]. Because the codebases for such systems are accessible to users and developers, custom modifications can easily be implemented, unlike proprietary systems. They may also offer seamless interoperation or integration options with minimal additional requirements for working with other open source tools. Another important feature of open source AutoML tools is the wide community involvement in the development and extension of functionalities. Generally, open source AutoML tools are more restricted in terms of the level of automation. They typically involve some level of manual work in the implementation of data processing tasks and model development. In particular, users are required to have some competence in the underlying programming languages (e.g., C++, Python or R) used to build the tools. It also requires a level of understanding of the modeling process – correct formulation of big data problem and an intuitive choice of possible models, as well as how to assess the suitability of resulting models. These tools are designed with flexibility and ease of adaptation and extension in mind. Compared to proprietary tools, they provide less intuitive graphical user interfaces (GUIs), and generally more difficult for non-expert users to processing data and building models for various tasks.

### 7.2.2   Proprietary AutoML tools

Some of the largest and most popular AutoML tools are proprietary solutions. Prominent among these include MS Azure AutoML [324], FLAML [309], H2O AutoML [100], Google Cloud AutoML [242], AutoGluon [315], IBM Watson AutoAI [327] and DataBricks [322]. Among these tools, one can also distinguish systems developed by technology start-ups (e.g., Darwin [326] by SparkCognition, OptiML [320] by BigML) and those owned by very big companies, so-called tech giants (e.g., AutoGluon [315] by Amazon and Google Cloud AutoML [242] by Alphabet). Solutions provided by start-ups are generally stand-alone products aimed at relatively narrow scope of applications. Those provided by technology giants tend to be more generic and focus on providing a broad range of tools to meet diverse business needs. The tools developed by tech giants are typically cloud-based platforms capable of handling large amounts of data for businesses as well as individual users. Prominent among these category of AutoML tools include Amazon SageMaker Autopilot [313], Google Cloud AutoML [242], Google Vertex AI [319], Microsoft Azure ML [324], IBM Watson AutoAI [327]. The tools provide easy-to-use graphical user interfaces and a rich set of tools within advanced integrated development environments that support complex tasks and processes. The cloud-based platforms are generally highly scalable and easily allow integration of other cloud-based utilities. Microsoft Azure ML Microsoft Azure ML [324], for instance, provides a suit of cloud-based AutoML tools such as Machine Learning API Service, Microsoft Azure Custom Vision and the Machine Learning Studio. In addition to providing utilities and APIs on the cloud that can be leverage by non-expert users to process training data and build machine learning models for various tasks, it allows one to seamlessly work directly with other Microsoft tools such as Microsoft Azure HDInsight and MS SQL Server. Similarly, Google Cloud AutoML provides modules such as AutoML Vision, AutoML Data Science and AutoML Natural Language that perform computer vision, Big Data processing and analytics, and NLP tasks, respectively. These toolsets are cloud-enabled, and can further integrate and interact with other Google products through the cloud.

Proprietary frameworks are designed to particularly target business users with subject matter knowledge about their technical job areas or business operations but without data science expertise. Since such users cannot easily understand what data processing tasks need to be performed, or create appropriate queries to explore and analyze data, this type of AutoML tools provide typically intuitive graphic user interfaces and visualization utilities to allow users to perform these data processing, exploration and analysis functions at a higher level.

## 8   IMPLICATIONS FOR INDUSTRY AND COMMERCE

The techniques discussed in this survey seek to automate mundane machine learning tasks and reduce errors associated with manual data processing procedures. Specifically, they eliminate all stages of data preparation and manual model selections, tuning and evaluation. This has led to greater productivity and improved accuracy of results in many areas. The automation of low-level data processing and model development steps allows business users to move straight to generating results from their raw input data. This reduces the time required for products and services to move from development to market. The enormous power of automated AI technologies also helps industries to process extremely large volumes of heterogeneous data. This allows them to better discover and exploit new insights to accelerate innovation, leading to the development of new products, services and even entirely new business models.

The transformative role of AI in business and industry is undisputed [338]. However, implementing practical AI solutions for industry and business requires highly qualified professionals with strong expertise in Statistics, Mathematics and Computer Science. Unfortunately, there is a severe shortage of qualified personnel [339], [340]. State-of-the-art AutoML and generative AI tools have simplified the development and application of AI solutions by encapsulating all complex algorithmic implementations behind intuitive user interfaces which support high-level interaction. The technologies have enabled the development of powerful AI tools and platforms that eliminate the need for AI specialists and allow non-expert users to be able to effortlessly perform otherwise complex data processing and machine learning tasks. They effectively eliminate the need for firms to hire expensive and scarce personnel. In practice, this means lowering the entry barrier for businesses, and thus empowering small businesses and industries with powerful AI capabilities. The immergence of cloud-based tools further expands the potential of businesses without the needed physical infrastructure to leverage AI capabilities for productivity. The simplification and significant reduction of cost of new models will undoubtedly expand the scope of AI and machine learning technologies in business and industry. We discuss the implications of these technologies for specific industries, namely: healthcare, agriculture, manufacturing, and retail, banking and finance.

### 8.1 Healthcare

The implications of automated data processing and the emergence of predictive analytical tools in healthcare are enormous. With the ability to process large volumes of medical data with minimal human intervention and provide useful insights and recommendations, these technologies enable medical professionals to make accurate diagnoses and provide appropriate treatment plans. Specifically, automated AI tools with access to large volumes of medical data–e.g., from wearable medical devices and IoT sensors, electronic medical records, scans, clinical tests and genetic profiles–together with advanced scientific knowledge bases (e.g., in the form of generative AI-based models like LLMs) can quickly analyze any new medical case, provide high-level, human-understandable information about the diagnosis and treatment options. Indeed, specialized LLM-based medical tools such as Med-PaLM 2 [341] and MedAlpaca [342] are already capable of expert-level medical question answering, and in some cases outperforming medical professionals in medical licensing examination [341].

The use of advanced data processing tools greatly helps to eliminates human factors from medical care, minimizing the risk of costly errors that humans sometimes commit as a result of distraction and fatigue.

Automated AI solutions also serve as a catalyst for innovation in medical treatment. For instance, they are able to analyze existing treatment options and reconfigure them to work better or to fight new medical conditions. They are able to provide more suitable treatment plans for individuals with peculiar underlying physiological issues. These data-driven tools are also important for tasks like drug discovery as they help in the identification of new drug targets, analysis of their effectiveness and prediction of potential drug reactions and side effects. The use of these modern technologies significantly increases the precision and efficiency of medical care while at the same time significantly reducing the costs.

In addition to extending the capabilities and service quality of medical professionals, these technologies, through online conversational AI tools (e.g., chatbots), also offer patients the means to access personalized healthcare anytime and anywhere; individuals who experience symptoms can consult chatbots for personalized medical advice and direction. This capability is especially useful as supportive care system for elderly people. This also frees precious time for medical professionals, allowing them to focus on important medical problems instead of taking patients' medical complaints and providing consultancy services in non-critical situations.

### 8.2 Agriculture

In agriculture, these techniques will simplify the complex task of obtaining and processing heterogeneous data for various field variables and conditions. Additionally, farmers will be able to utilize advanced digital tools to automate complex agricultural tasks that would have otherwise required technical expertise in AI and machine learning. Specifically, they will be able to develop advanced predictive analytics models using AutoML and generative AI tools to predict and manage risks such as disease and pest infestation, changes in weather and climate patterns, and other factors for crop failure. Using these tools, lay farmers will be able forecast market demand for their produce and predict the expected crop yield, as well as provide better service to their suppliers, distributors and downstream customers.

### 8.3 Retail, banking and finance

Another important area of significant impact of automated data processing techniques is retail, banking and finance. Here, AI-enabled technologies have simplified and automated important business processes (e.g., processing and analysis of financial data, management related tasks, analysis of documents to aid compliance with regulatory requirements). They offer a cheap and an effective means to discover important business patterns, trends and opportunities while simultaneously providing useful and actionable directions to responsible personnel to respond appropriately. With the help of automated data processing techniques companies are able to focus on business problems instead of on data processing and analysis, thereby increasing their productivity and profitability. By processing customer data in advanced analytics engines, companies are also able to predict customer behavior and preference, and thus provide tailored products and services to better address their needs. Generative AI tools developed on the basis of business-specific data also enhance customer experience by automating customer engagement tasks in the form of advanced virtual assistants that can provide context-specific information and personalized financial recommendations to clients in real-time.

### 8.4 Manufacturing

In manufacturing, automated AI tools can also help to streamline overall factory operations as well as automate, optimize and manage energy consumption, logistics, and production. Data gathered from IoT devices and sensors can be analyzed by automated machine learning models to provide useful information about potential hazards and recommend preventive actions. Generative AI techniques allows manufacturing companies to design new products simply by specifying a set of desired attributes. This particular capability has a huge potential in manufacturing as it can lead to innovative product concepts and products with qualitatively better characteristics than their counterparts designed by traditional methods. For instance, such products may be more efficient, cheaper and environmentally-friendly.

## 9 DISCUSSIONS

In the era of big data, the volume and complexity of data that machine learning systems typically handle have increased substantially. Consequently, collecting and processing the data into a form that is suitable for machine learning tasks is a challenging undertaking. Approaches based on conventional machine learning concepts are extremely laborious and require enormous development time. Automated processing methods, especially approaches based on AutoML, have greatly automated these laborious processes and, thus, have simplified and accelerated the development cycle of deep learning models. As the size data and the complexity of machine learning problems increase, such approaches are expected to become a general practice, especially for generic machine learning tasks such as Big Data analytics and data visualization.

### 9.1 Main challenges

We discuss some of the most important challenges modern automated data processing systems have to deal with.

#### 9.1.1 Growing volume and complexity of data

While more and more data processing tasks are becoming increasingly automated, today's AutoML methods are still limited in terms of the complexity of data and tasks that can be handled. Handling very complex data still requires the intervention of human developers at some stages in the development process. While many of the recent works have focused primarily on reducing the amount of computational resources needed to implement automated data processing solutions, the techniques continue to evolve steadily and many new workarounds are expected be introduced in the foreseeable future to incrementally extend the scope of tasks that can be automated. Future AutoML systems will have the ability to create end solutions for complex problems by automating the full range of tasks from input data acquisition to model construction and validation.

Since many data processing tools are intended for use by different types of users and the resulting models typically need to meet several performance objectives simultaneously, albeit within an acceptable cost, it is often challenging to ensure that the right balance of performance and complexity is achieved. To address this issue, it is possible to develop a flexible data processing scheme which can be tweaked per the specific user or application requirements. For instance, Tsamardinos et al. in [57] devised different model configuration options, with each configuration prioritizing a specific machine learning objective: interpretability, predictive performance, minimization of data size. These configuration settings allow users to customize model performance according to their priorities. This approach is used by data processing engines in many general-purpose automated machine learning systems that handle large-scale data. However, these state-of-the-art approaches still require users to have an understanding of the underlying requirements and to manually select the best settings for the particular task. Future work could employ context knowledge to automate this process, too.

#### 9.1.2 Complexity of data processing problems

The application of machine learning in more diverse application areas, coupled with the rapidly increasing complexity of data pipelines has made it challenging to solve many modern problems using automated machine learning. Preprocessing tasks such as labeling and categorical data annotation are particularly difficult to automate. Because of these challenges, in practical applications, AutoML systems usually automate only some data processing tasks in the machine learning pipeline and provide baseline results that can reveal further insights into possible avenues for improvement by manual means. In the near future, the introduction of a wide variety of new data processing techniques and algorithms for their optimization will provide a means for solving complex problems whose solutions are unattainable at present.

#### 9.1.3 Context-awareness

Approaches based of state-of-the-art AutoML often generate a large pool of intermediate features and attributes in the search space. The performance of the end model ultimately, to a large extent, depends on the search space. Because the generation process is inherently "blind", generated features might not be informative or relevant to the target task or could introduce redundant information. This could, even with good optimization methods, lead to poor performance.

#### 9.1.4 Overly conservative results

To avoid making bizarre errors, generative models are typically designed to be very conservative and tend to produce outputs that align well within the distribution of the encoded training data, thus avoiding extreme cases and outliers. In many real-world domains, however, data is often not clean. Consequently, the output data of generative models often lack the wide variability of real-world data. For instance, large language models generally fail to produce context-specific responses and resort to overly generic answers. Training with such data may result in less-than-acceptable outcomes in some cases. Moreover, because of the enormous sizes, complexity and opacity of generative AI-based automated data processing and synthesis techniques such as diffusion models and LLMs, they can sometimes produce unreliable and potentially harmful data that may be difficult to detect. The approaches are also prone to the so-called hallucination effect [343]. These problems severely limit the utility of generative AI models in high-stakes tasks, especially in application domains such as finance, security and healthcare.

### 9.1.5    Adaptability and transferability

Because the processing procedures are not based on intuitive and well-grounded mechanisms, data generated by automated techniques might be specifically optimized for a singlr task or a narrow set of tasks. This may produce good results for the specific models and datasets they have been tuned for but fail to generalize well in unseen data and new problem domains. The problem of transferablility and adaptability of automatically generated data has not been investigated in the literature. Also, as data in many application settings (e.g., data about pandemics, economic or market conditions) evolve over time, automated processing methods need to be able to adapt accordingly. Moreover, it is often useful to have a single solution that can address a set of related problems in a broad application domain. Current approaches are not able to meet this  need.

### 9.1.6    Balancing multiple requirements and trade-offs

Modern automated data processing problems often involve very large datasets with complex relationships and interactions. At the same time, models that are trained on them are required to meet multiple  performance requirements. In particular, there is the need to navigate delicate trade-offs among different aspects of performance, such as reducing data size versus retaining important information, achieving higher predictive accuracy versus interpretability, robustness and algorithmic fairness. Methods that that enhance performance in one aspect might harm performance in another. These complex trade-offs make it difficult for automated machine learning algorithms to strike the right balance independent of human input, as users may prioritize different aspects of performance according to their specific goals.

### 9.1.7    Reliability and trust

With the widespread adoption of AI in diverse areas, their reliability – and consequently trust in their solutions –has become an important issue. This is particularly more serious with automated learning methods since they rely on black-box approaches automated hyperparameter optimization and there is very little human supervision of the overall learning process. The automated optimization of hyperparameters might result in complex settings and potentially harmful interactions that are difficult to predict or diagnose at design time but may manifest later under deployment. Moreover, with these black box methods, it is challenging to guarantee that the end systems will perform as intended and align solutions with broader user objectives while respecting societal values and expectations– i.e., make accurate inferences while at the same time being constrained by wider issues such as safety, privacy and ethics. It is currently not possible to incorporate these high-level concepts in the automated learning process.

### 9.1.8    Lack of comprehensive valuation metrics for some processing tasks

Unlike many machine learning tasks where standard model configurations, datasets and evaluation metrics have been created to test different methods, for many data processing tasks, especially preprocessing, there is a general lack of standardized settings and metrics that can be universally applied across different tasks and datasets. This leads to researchers using varying settings and metrics, thereby making direct comparison of methods very challenging.

### 9.1.9    Scalability

Scalability is another major limitation of automated data processing methods. White these techniques work well for large-scale problems involving huge datasets, they typically perform very poorly on small data; training effective when large datasets used. In situations where data is scarce, traditional processing techniques often outperform automated methods. Furthermore, automated processing methods are inherently computationally expensive. Therefore, while automated data processing approaches are promising, the choice between automated methods and traditional approaches ultimately depends on factors such as dataset size, problem domain, and available computational resources.

### 9.1.10    Limited scope of application

Existing AutoML tools are limited in terms of the range of tasks and data types they support. Presently, most generic solutions work best for tabular data. Also, most automated data processing techniques support classification and regression problems, while offering little in problem domains such as natural language processing and time series forecasting.

## 9.2    Future prospects

Since the concept of automated data processing based on AutoML methods is fairly new, one would assume that we are just at the initial stage of realizing the vast potential this approach presents. At the same time, it is important to exercise caution when making assessments about its future possibilities as undue expectations can ultimately lead to disappointment, which can potentially result in "automated data processing winter" [344]. Also, because automated data acquisition and processing methods are characteristically resource intensive, their potential will depend on the progress in other areas such as computer hardware technology and the development and extensive use of cloud computing infrastructure. The main prospects for the foreseeable future include the following.

### 9.2.1 Support for higher-level functions

New data preprocessing algorithms that also perform quality control and budget management functions taking into consideration the specific application requirements as well as the objectives and priorities of the end user will be developed. This will involve more advanced user interface (UI) and user experience (UX) solutions with intuitive features that support high-level semantic interaction.

### 9.2.2 Extension of generic AutoML methods to handle more complex problems and diverse data types

While specialized automated data processing methods have partially tackled a variety of data types and machine learning problems, presently, generic AutoML solutions are designed primarily to work with tabular data; their support for other data types such as images, audio, video and point clouds is very limited. Also, they mostly solve general classification and regression tasks, and poorly handle data processing for applications in time series and natural language processing domains. In the near future, new approaches that extend the scope of automated data processing methods, especially those based on AutoML frameworks, to a wider problem domain, including time series and natural language processing, are envisaged.

### 9.2.3 Advanced human-in-the-loop (HITL) automated data processing

Future research will concentrate on developing methods to provide user-friendly interfaces and more intuitive explanations for human actors based on techniques such as feature importance visualization, counterfactual explanations, and attention mechanisms. In turn, researchers will develop machine learning algorithms that can learn from human feedback through, for example, explicit labeling and reinforcement signals. Together, these two sets of approaches can be leveraged to provide synergy between humans and machines. This will ultimately lead to improved performance, reliability, robustness, and the ability to tackle more challenging problems that are beyond the capability of either humans or machines when working independently. It is conceivable that more advanced HITL systems will enable support for interaction of multiple users or user-expert groups with automated machine learning systems. Businesses, for example, will be able to model high-level information such as broad organizational goals and priorities, available budgetary resources, and the overall bottom line. Systems based on this approach will also align better with broader human needs, values, and expectations.

Humans in the machine learning loop will be particularly useful for more complex data processing tasks such as categorical encoding and data annotation. Humans will also play a vital role in model validation and refinement. as well as the incorporation of high-level concepts such as

### 9.2.4 New and dedicated infrastructure for automated data processing

Dedicated tools that specifically cater to the needs of big and complex data problems will be developed. These will be specialized platforms (e.g., cloud-based infrastructure), tools (e.g., libraries and open-source software), and frameworks (e.g., generic data processing models) designed to support and streamline different aspects of automated data processing. This infrastructure will simplify various processing tasks and make it easier for both data scientists and non-experts to effectively use automated machine learning models for data processing. This type of infrastructure will seamlessly provide needed functions while taking care of contemporary challenges such as privacy, transparency, and data security.

### 9.2.5 Progressive self-refinement of synthetic data

It is important to note that while automated data processing models can perform complex data processing and can even synthesize relevant data automatically, paradoxically, their training requires large volumes of high-quality data that correctly captures the distribution of data in the target domains. For many domains data are not often available naturally. However, it is conceivable that in the future, in cases where domain-relevant data may be inaccessible, more generic generative AI models could be used to synthesize rudimentary data and then perform progressive self-refinement by continually generating, filtering and reusing the generated data for subsequent retraining and fine-tuning until an acceptable quality of data is attained. This approach is already being studied [345], and preliminary results are promising.

### 9.2.6 Explainable data synthesis by generative AI techniques

In the near future, researchers will focus on developing models that not only deliver high performance but also provide insights for their actions. Explainable AutoML frameworks designed for data processing will be able to perform data processing tasks in an end-to-end machine learning pipeline while providing explanations about the intermediate processes and resulting outcomes. This approach will be particularly useful for complex data processing tasks such as categorical encoding and data annotation. Explainability will also play a vital role in allowing researchers to better evaluate and validate results, as well as to perform further refinements.

Generative AI techniques will be particularly useful in solving the problem of opacity associated with the data generation process of traditional AutoML methods. For instance, the massive volume of knowledge embedded in LLMs and diffusion models can be leveraged to explain the underlying AutoML model's decisions regarding generated data. Specifically, these models will be able to create additional metadata about the generated data, including information about the relationships among different data elements and attributes. In addition, intuitive user interfaces and high-level interaction mechanisms will allow developers to incorporate human-understandable information about the target tasks for which the dataset is to be generated. This can help to mitigate inadequacies and potential flaws in the data generated by today's generative AI methods.

## 10 CONCLUSION

The importance of automated data processing has increased remarkably in the last few years. This is largely driven by the increasing demand for machine learning solutions in many areas, coupled with the large volumes of data that need to be processed for these machine learning tasks.

In this work, we survey state-of-the-art approaches for automating data processing for deep learning and big data tasks. We first present methods for realizing individual data processing solutions. These include data preprocessing (e.g., data cleaning, imputation, labeling, categorical encoding, etc.), data augmentation and feature engineering (specifically, feature extraction, construction and selection). We also discuss approaches to implementing all processing steps holistically within a single end-to-end deep learning framework. We summarized the main characteristics and functions of generic AutoML frameworks designed for big data applications. Furthermore, we discuss future developments that are likely to have a significant impact on the success of automated data processing.

The survey shows that while many data processing tasks can already be seamlessly automated in state-of-the-art automated machine learning pipelines, challenges still remain regarding the full automation certain tasks. The need to address a wide scope of problems makes it particularly difficult for machine learning systems to incorporate effective search mechanisms that allow relevant data to be collected and exploited in a context-dependent manner.

## Acknowledgments

The authors would like to thank...